\documentclass[12, a4paper]{article}
\usepackage[utf8]{inputenc}

\usepackage[english]{babel}

\usepackage[explicit]{titlesec}
\usepackage{blindtext}

\usepackage{lineno}
\usepackage{geometry}
\usepackage{setspace}
\usepackage{indentfirst}
\usepackage{csquotes}

\usepackage[hidelinks]{hyperref}
\usepackage[backend=biber, style=apa, sorting=nyt, uniquelist=false, dashed=false, bibencoding=utf8]{biblatex}
\usepackage{fancyhdr}
\usepackage{graphicx}
\usepackage{float}
\usepackage[center]{caption}
\usepackage{subcaption}
\usepackage{array}
\usepackage{makecell}

\usepackage{tabularx,ragged2e,booktabs,caption,tabu}
\newcolumntype{L}[1]{>{\raggedright\arraybackslash}p{#1}}
\usepackage{longtable}
\usepackage{multirow}
\usepackage[table,xcdraw]{xcolor}
\newcolumntype{P}[1]{>{\centering\arraybackslash}p{#1}}
\newcolumntype{M}[1]{>{\centering\arraybackslash}m{#1}}
\newcolumntype{Y}[1]{>{\centering\arraybackslash}X{#1}}
\usepackage{placeins}
\usepackage{rotating}
\usepackage{lscape}
\usepackage{scrhack}
\usepackage{supertabular}
\usepackage{afterpage}

\usepackage{authblk} 
\usepackage{pgfplots, pgfplotstable}
\usepackage{tikz}
\usetikzlibrary{shapes.geometric, arrows}
\usetikzlibrary{mindmap,trees,positioning}
\usepackage{smartdiagram}
\usesmartdiagramlibrary{additions}
\usepackage{verbatim}
\tikzstyle{bag} = [align=center]
\pgfplotsset{compat=1.9}
\usetikzlibrary{arrows, decorations.markings}

\begin{document}

\title{A Survey of Deep Learning Techniques for Weed Detection from Images}
\def\correspondingauthor{\footnote{Corresponding author email: 33916214@student.murdoch.edu.au}}
\author[1,2]{A S M Mahmudul Hasan\correspondingauthor{}}
\author[1,2]{Ferdous Sohel}
\author[2,3,4]{Dean Diepeveen}
\author[1,5]{Hamid Laga}
\author[2] {Michael G.K. Jones}
\affil[1]{Information Technology, Murdoch University, Murdoch, WA 6150, Australia}
\affil[2]{Centre for Crop and Food Innovation, Food Futures Institute, Murdoch University, Murdoch, WA 6150, Australia}
\affil[3]{Department of Primary Industries and Regional Development, Western Australia, South Perth, WA, 6151, Australia}
\affil[4]{Centre for Sustainable Farming Systems, Murdoch University, Murdoch, WA 6150, Australia}
\affil[5]{Centre of Biosecurity and One Health, Harry Butler Institute, Murdoch University, Murdoch University, Murdoch, WA 6150, Australia}

\date{\vspace{-5ex}}

\maketitle

\pagenumbering{arabic}


\begin{abstract}
The rapid advances in Deep Learning (DL) techniques have enabled rapid detection, localisation, and recognition of objects from images or videos.  DL techniques are now being used in many applications related to agriculture and farming. Automatic detection and classification of weeds can play an important role in weed management and so contribute to higher yields. Weed detection in crops from imagery is inherently a challenging problem because both weeds and crops have similar colours (‘green-on-green’), and their shapes and texture can be very similar at the growth phase. Also, a crop in one setting can be considered a weed in another. In addition to their detection, the recognition of specific weed species is essential so that targeted controlling mechanisms (e.g. appropriate herbicides and correct doses) can be applied. In this paper, we review existing deep learning-based weed detection and classification techniques. We cover the detailed literature on four main procedures, i.e., data acquisition, dataset preparation, DL techniques employed for detection, location and classification of weeds in crops, and evaluation metrics approaches. We found that most studies applied supervised learning techniques, they achieved high classification accuracy by fine-tuning pre-trained models on any plant dataset, and past experiments have already achieved high accuracy when a large amount of labelled data is available.
\end{abstract}


\section{Introduction}

The world population has been increasing rapidly, and it is expected to reach nine billion by 2050. Agricultural production needs to increase by about 70\% to meet the anticipated demands  \parencite{radoglou2020compilation}. However, the agricultural sector will face many challenges during this time, including a reduction of cultivatable land and the need for more intensive production. Other issues, such as climate change and water scarcity, will also affect productivity. Precision agriculture or digital agriculture can provide strategies to mitigate these issues  \parencite{lal1991soil,seelan2003remote,radoglou2020compilation}.\par

Weeds are plants that can spread quickly and undesirably, and can impact on crop yields and quality  \parencite{patel2016weed}. Weeds compete with crops for nutrition, water, sunlight, and growing space \parencite{iqbal2019investigation}. Therefore, farmers have to deploy resources to reduce weeds. The management strategies used to reduce the impact of weeds depend on many factors. These strategies can be categorised into five main types \parencite{sakyi_2019}: ‘preventative’ (prevent weeds from becoming established), ‘cultural’ (by maintaining field hygiene – low weed seed bank), ‘mechanical’ (e.g., mowing,  mulching and tilling), ‘biological’ (using natural enemies of weeds such as insects, grazing animals or disease), and ‘chemical’ (application of herbicides). These approaches all have drawbacks. In general, there is a financial burden and they require time and extra work. In addition, control treatments may impact the health of people, plants, soil, animals, or the environment  \parencite{okese_kankam_boamah_evans_2020,sakyi_2019, holt2004principles}.\par

As the costs of labour has increased, and people have become more concerned about health and environmental issues, automation of weed control has become desirable \parencite{liu2020weed}. Automated weed control systems can be beneficial both economically and environmentally. Such systems can reduce labour costs by using a machine to remove weeds and, selective spraying techniques can minimise the use of the herbicides \parencite{lameski2018review}.\par 

\begin{figure}[tb]
    \centering
    \captionsetup[subfigure]{justification=centering}
    \begin{subfigure}[t]{.3\textwidth}
        \centering
        \includegraphics[width=\textwidth, height = 10em]{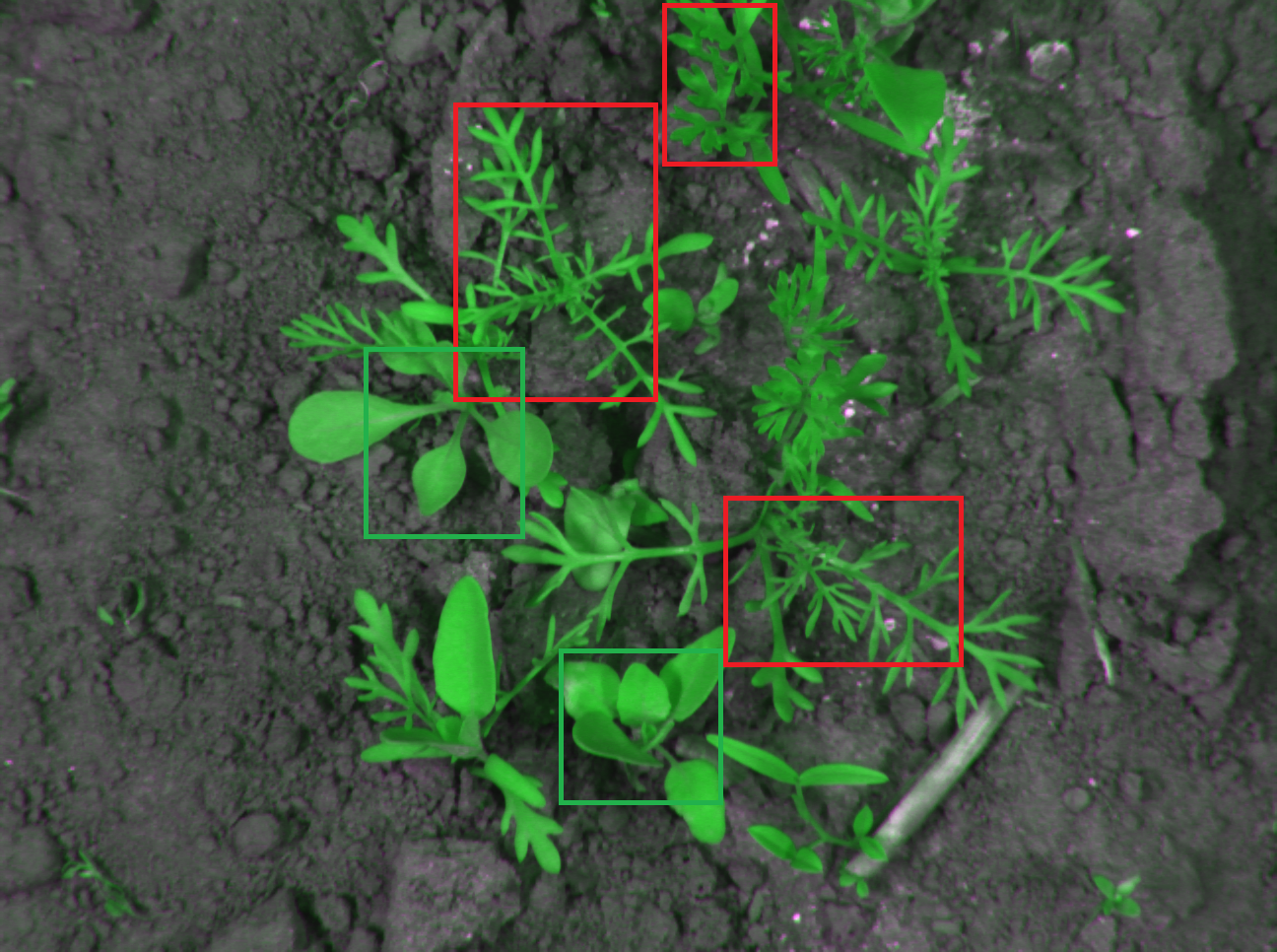}  
        \caption{Occlusion of crop and weed \parencite{haug2014crop}}
        \label{fig:cw}
    \end{subfigure}
    \hfill
    \begin{subfigure}[t]{.3\textwidth}
        \centering
        \includegraphics[width=\textwidth, height = 10em]{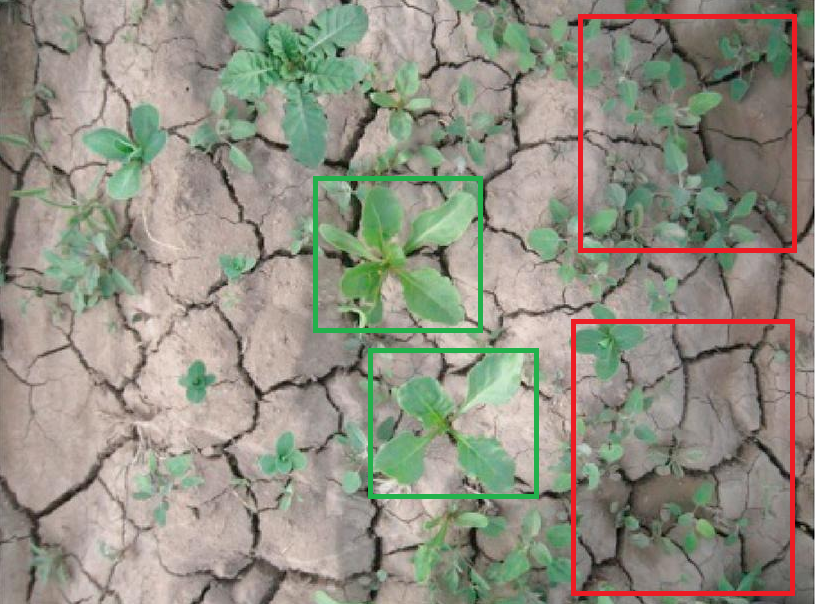}  
        \caption{Colour and texture similarities between crop and weed plants \parencite{bakhshipour2018evaluation}}
        \label{fig:sw}
    \end{subfigure}
    \hfill
    \begin{subfigure}[t]{.3\textwidth}
        \centering
        \includegraphics[width=\textwidth, height=10em]{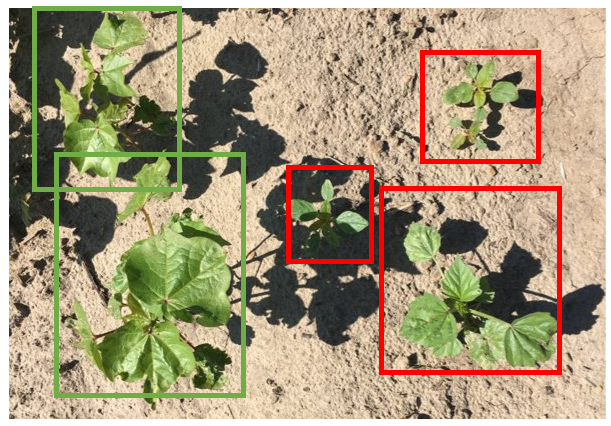}  
        \caption{Shadow effects in natural weed image \parencite{pytorch_2020}}
        \label{fig:cpw}
    \end{subfigure}
    \hfill
    \begin{subfigure}[t]{.3\textwidth}
        \centering
        \includegraphics[width=\textwidth, height = 10em]{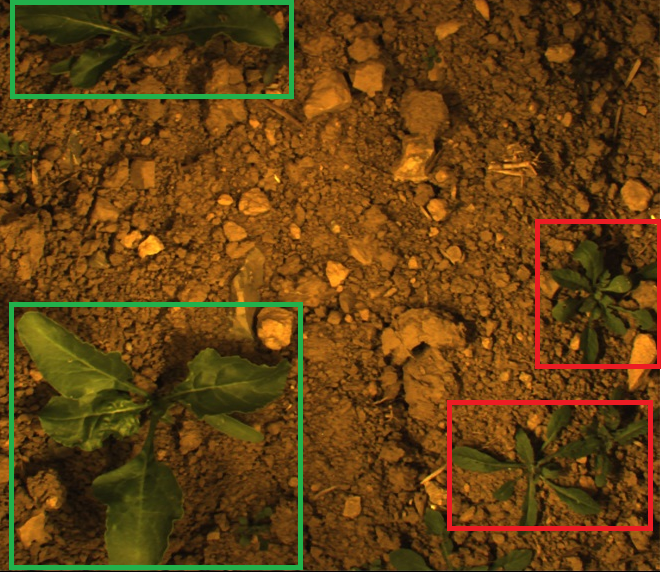}  
        \caption{Effects of illumination conditions \parencite{di2017automatic}}
        \label{fig:sbw}
    \end{subfigure}
    \hfill
    \begin{subfigure}[t]{.3\textwidth}
        \centering
        \includegraphics[width=\textwidth, height = 10em]{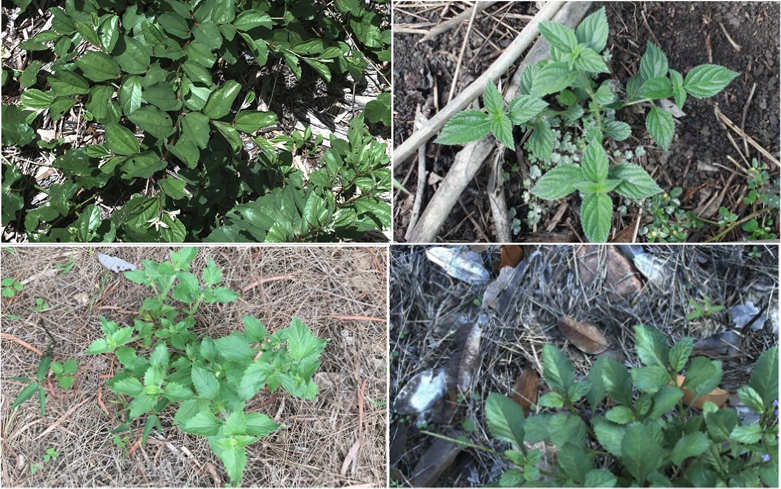}  
        \caption{Four different species of weeds that share similarities (inter-class similarity) \parencite{olsen2019deepweeds}}
        \label{fig:fws}
    \end{subfigure}
    \hfill
    \begin{subfigure}[t]{.3\textwidth}
        \centering
        \includegraphics[width=\textwidth, height = 10em]{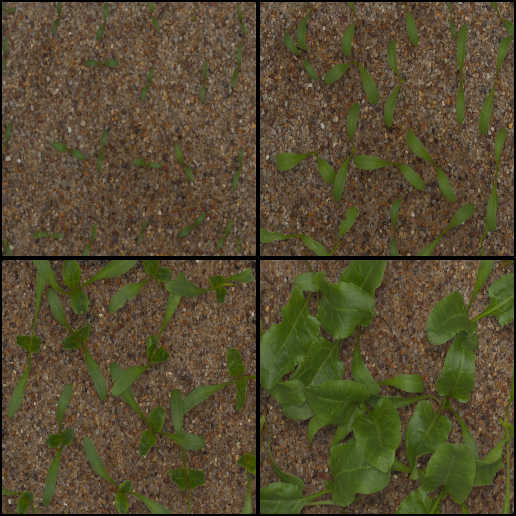}  
        \caption{Sugar beet crop at different growth stages (intra-class variations) \parencite{giselsson2017public}}
        \label{fig:gs}
    \end{subfigure}
    \hfill
    \begin{subfigure}[t]{.3\textwidth}
        \centering
        \includegraphics[width=\textwidth, height = 10em]{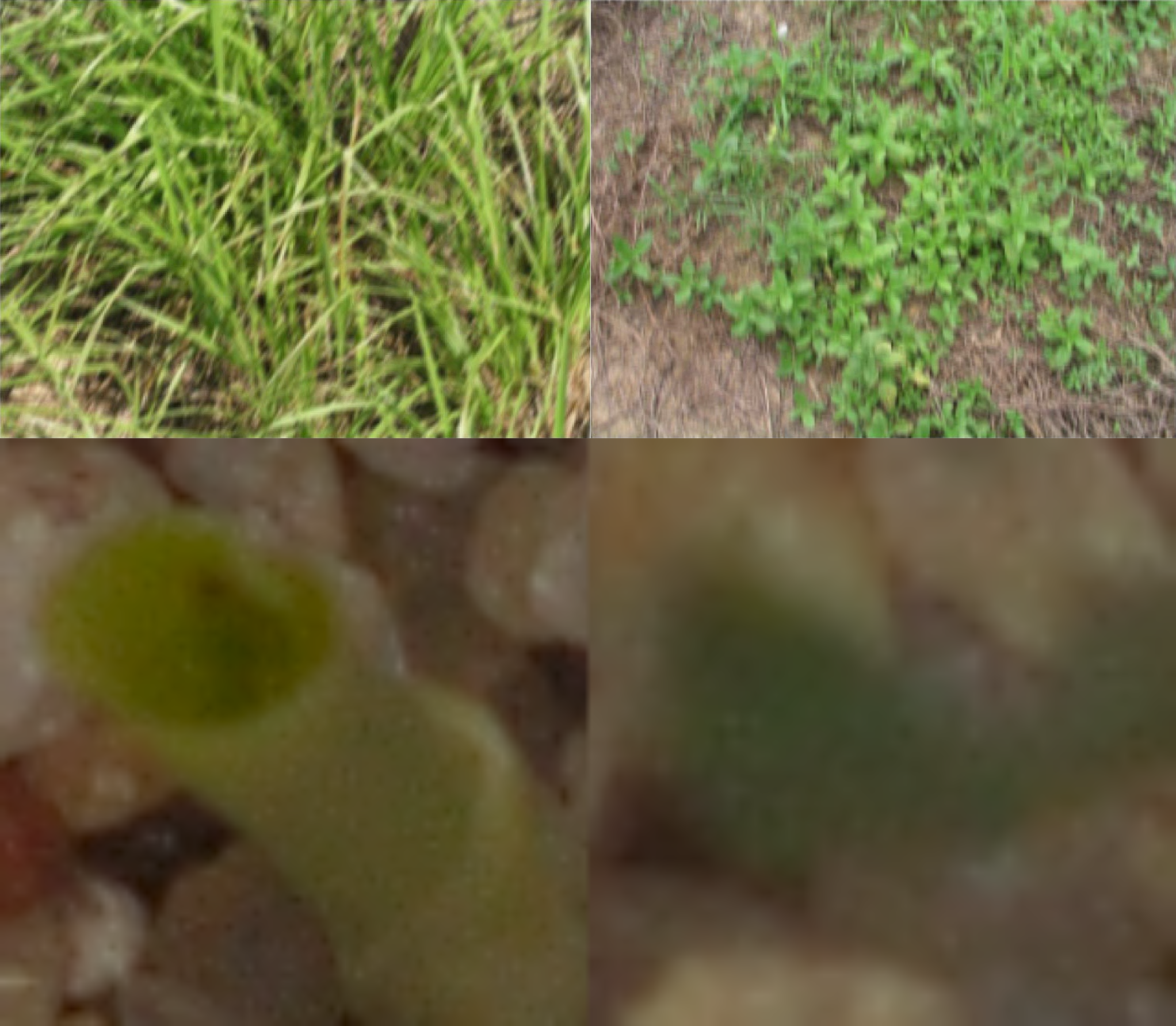}  
        \caption{Effects of motion blur and noise \parencite{giselsson2017public, ahmad2018visual}}
        \label{fig:bn}
    \end{subfigure}
    \hfill
    \begin{subfigure}[t]{.65\textwidth}
        \centering
        \includegraphics[width=\textwidth, height = 10em]{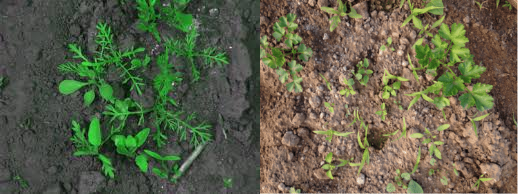}  
        \caption{Weeds can vary at different geographic/weather locations: weed in carrot crop collected from Germany(left) \parencite{haug2014crop} and Macedonia (Right) \parencite{lameski2017weed}}
        \label{fig:gl}
    \end{subfigure}

\caption[Variations and challenges in image based weeds detection (green boxes indicate crops and red boxes indicate weeds).]{Weeds in different crops (green boxes indicate crops and red boxes indicate weeds).}
\label{fig:weeds_in_crops}
\end{figure}

To develop an automatic weed management system, an essential first step is to be able to detect and recognise weeds correctly \parencite{liu2020weed}. Detection of weeds in crops is challenging as weeds and crop plants often have similar colours, textures, and shapes. Figure \ref{fig:weeds_in_crops} shows crop plants with weeds growing amongst them. Common challenges in detection and classification of crops and weeds are occlusion (Figure \ref{fig:cw}), similarity in colour and texture (Figure \ref{fig:sw}), plants shadowed in natural light (Figure \ref{fig:cpw}), colour and texture variations due to lighting conditions and illumination (Figure \ref{fig:sbw}) and different species of weeds which appear similar (Figure \ref{fig:fws}). Same crop plants or weeds may show dissimilarities during growth phases (Figure \ref{fig:gs}). Motion blur and noise in the image also increase the difficulty in classifying plants (Figure \ref{fig:bn}). In addition, depending on the geographical location (Figure \ref{fig:gl}) and the variety of the crop, weather and soil conditions, the species of weeds can vary \parencite{jensenautomated}.\par

A typical weed detection system follows four key steps: image acquisition, pre-processing of images, extraction of features and detection and classification of weeds \parencite{shanmugam2020automated}. Different emerging technologies have been used to accomplish these steps. The most crucial part of these steps is weed detection and classification. In recent years, with advances in computer technologies, particularly in graphical processing units (GPU), embedded processors coupled with the use of Machine Learning (ML) techniques have become more widely used for automatic detection of weed species \parencite{lecun2015deep, gu2018recent, yu2019deep}.\par

Deep learning (DL) is an important branch of ML. For image classification, object detection, and recognition, DL algorithms have many advantages over traditional ML approaches (in this paper, the term machine learning, we mean traditional machine learning approaches). Extracting and selecting discriminating features with ML methods is difficult because crops and weeds can be similar. This problem can be addressed efficiently by using DL approaches based on their strong feature learning capabilities. Recently, many research articles have been published on DL-based weed recognition, yet few review articles have been published on this topic. Su \parencite{su2020advanced} recently published a review paper in which the main focus was on the use of point spectroscopy, RGB, and hyperspectral imaging to classify weeds in crops automatically. However, most of the articles covered in this review have applied traditional machine learning approaches, with few citations of recent papers. \textcite{liu2020weed} analysed a number of publications on weed detection, but from the perspective of selective spraying. \par

We provide this comprehensive literature survey to highlight the great potential now presented by different DL techniques for detecting, localising, and classifying weeds in crops. We present a taxonomy of the DL techniques for weed detection and recognition, and classify major publications based on that taxonomy. We also cover data collection, data preparation, and data representation approaches. We provide an overview of different evaluation metrics used to benchmark the performance of the techniques surveyed in this article. \par

The rest of the paper is organised as follows. Existing review papers in this area are discussed briefly  in Section \ref{relatedWorks}. Advantages of DL-based weed detection approaches over traditional ML methods are discussed in Section \ref{ML_vs_DL}. In Section \ref{paper_selection}, we describe how the papers for review were selected. A taxonomy and an overview of DL-based weed detection techniques are provided in Section \ref{overview}. We describe four major steps of DL-based approaches, i.e. data acquisition (Section \ref{Data_Acquisition}), dataset preparation (Section \ref{dataset_preparation}), detection and classification methods (Section \ref{Deep_Learning_Architecture}) and evaluation metrics (Section \ref{Evaluation_Matrics}). In Section \ref{Detection_Approach} we have highlighted the approaches to detection of weeds in crop plants adopted in the related work. The learning methods applied the relevant studies are explained in Section \ref{learning_mehod}. We summarise the current state in this field and provide future directions in Section \ref{discussion} with conclusions are provided in Section \ref{conclusion}.\par

\section{Related Surveys} \label{relatedWorks}

ML and DL techniques have been used for weed detection, recognition and thus for weed management. In 2018, \textcite{kamilaris2018deep} published a survey of 40 research papers that applied DL-techniques to address various agricultural problems, including weed detection. The study reported that DL-techniques outperformed more than traditional image processing methods.\par

In 2016, \textcite{Merfield2016} discussed ten components that are essential and possible obstructions to develop a fully autonomous mechanical weed management system. With the advance in DL, it seems that the problems raised can now be addressed. \textcite{amend2019weed} articulated that DL-based plant classification modules can be deployed not only in weed management systems but also for fertilisation, irrigation, and phenotyping. Their study explained how \enquote{Deepfield Robotics} systems could reduce labour required for weed control in agriculture and horticulture.\par      

\textcite{wang2019review} highlighted that the most challenging part of a weed detection techniques is to distinguish between weed and crop species. They focused on different machine vision and image processing techniques used for ground-based weed detection. \textcite{brown2005site} made a similar observation. They reviewed remote sensing for weed mapping and ground-based detection techniques. They also reported the limitations of using either spectral or spatial features to identify weeds in crops.  According to their study, it is preferable to use both features.\par

\textcite{fernandez2018current} reviewed technologies that can be used to monitor weeds in crops. They explored different remotely sensed and ground-based weed monitoring systems in agricultural fields. They reported that weed monitoring is essential for weed management. They foresaw that the data collected using different sensors could be stored in cloud systems for timely use in relevant contexts. In another study, \textcite{moazzam2019review} evaluated a small number of DL approaches used for detecting weeds in crops. They identified research gaps, e.g., the lack of large crop-weed datasets, acceptable classification accuracy and lack of generalised models for detecting different crop plants and weed species. However, the article only covered a handful of publications and as such the paper was not thorough and did not adequately cover the breadth and depth of the literature.\par

\section{Traditional ML- vs DL-based Weed Detection Methods} \label{ML_vs_DL}
A typical ML-based weed classification technique follows five key steps: image acquisition, pre-processing such as image enhancement, feature extraction or with feature selection, applying an ML-based classifier and evaluation of the performance \parencite{liu2020weed,cesar2020comparison, bini2020machine,liakos2018machine}. 

Different image processing methods have been applied for crop and weed classification \parencite{woebbecke1995shape, hemming2002image, tian2000machine}. By extracting shape features, many researchers identify weeds and crops using discriminate analysis \parencite{chaisattapagon1995effective, meyer1998textural}. In some other research, different colour \parencite{zheng2017maize, jafari2006weed, hamuda2017automatic, kazmi2015detecting} and texture \parencite{bakhshipour2017weed} features were used. \par

The main challenge in weed detection and classification is that both weeds and crops can have very similar colours or textures. Machine learning approaches learn the features from the training data that are available \parencite{bakhshipour2018evaluation}. Understandably, for traditional ML-approaches, the combination of multiple modalities of data e.g. the shape, texture and colour or a combination of multiple sensor data is expected to generate superior results to a single modality of data. \textcite{kodagoda2008weed} argued that colour or texture features of an image alone are not adequate to classify wheat from weed species Bidens pilosa. They used Near-Infrared (NIR) image cues with those features. \textcite{sabzi2020automatic} extracted eight texture features based on the grey level co-occurrence matrix (GLCM), two spectral descriptors of texture, thirteen different colour features, five moment-invariant features, and eight shape features. They compared the performance of several algorithms, such as the ant colony algorithm, simulated annealing method, and genetic algorithm for selecting more discriminative features. The performance of the Cultural Algorithm, Linear Discriminant Analysis (LDA), Support Vector Machine (SVM), and Random Forest classifiers were also evaluated to distinguish between crops and weeds.\par

\textcite{karimi2006application} applied SVM for detecting weeds in corn from hyperspectral images. In other research, \textcite{wendel2016self} used SVM and LDA for classifying plants. They proposed a self-supervised approach for discrimination. Before training the models, they applied vegetation separation techniques to remove background and different spectral pre-processing to extract features using Principal Component Analysis (PCA). \textcite{ishak2007weed} extracted different shape features and the feature vectors were evaluated using a single-layer perceptron classifier to distinguish narrow and broad-leafed weeds. \par

\tikzstyle{startstop} = [rectangle, rounded corners, minimum width=3cm, minimum height=1cm,text centered, draw=black, fill=red!30]
\tikzstyle{process} = [rectangle, minimum width=3cm, minimum height=1cm, text centered, draw=black, fill=orange!30]
\tikzstyle{decision} = [diamond, minimum width=3cm, minimum height=1cm, text centered, draw=black, fill=green!30]
\tikzstyle{elli}=[draw, ellipse, fill = green!20,text centered, text width = 5em]
\tikzstyle{elli2}=[draw, ellipse, fill = green!20,text centered, text width = 15em]
\tikzstyle{block} = [draw, rectangle,rounded corners, fill = red!20, text width = 10em, minimum height = 10mm, text centered]
\tikzstyle{arrow} = [draw, -latex']
\tikzstyle{vecArrow} = [thick, decoration={markings,mark=at position
   1 with {\arrow[semithick]{open triangle 60}}},
   double distance=1.4pt, shorten >= 6pt,
   preaction = {decorate},
   postaction = {draw,line width=2pt, white,shorten >= 4.5pt}]

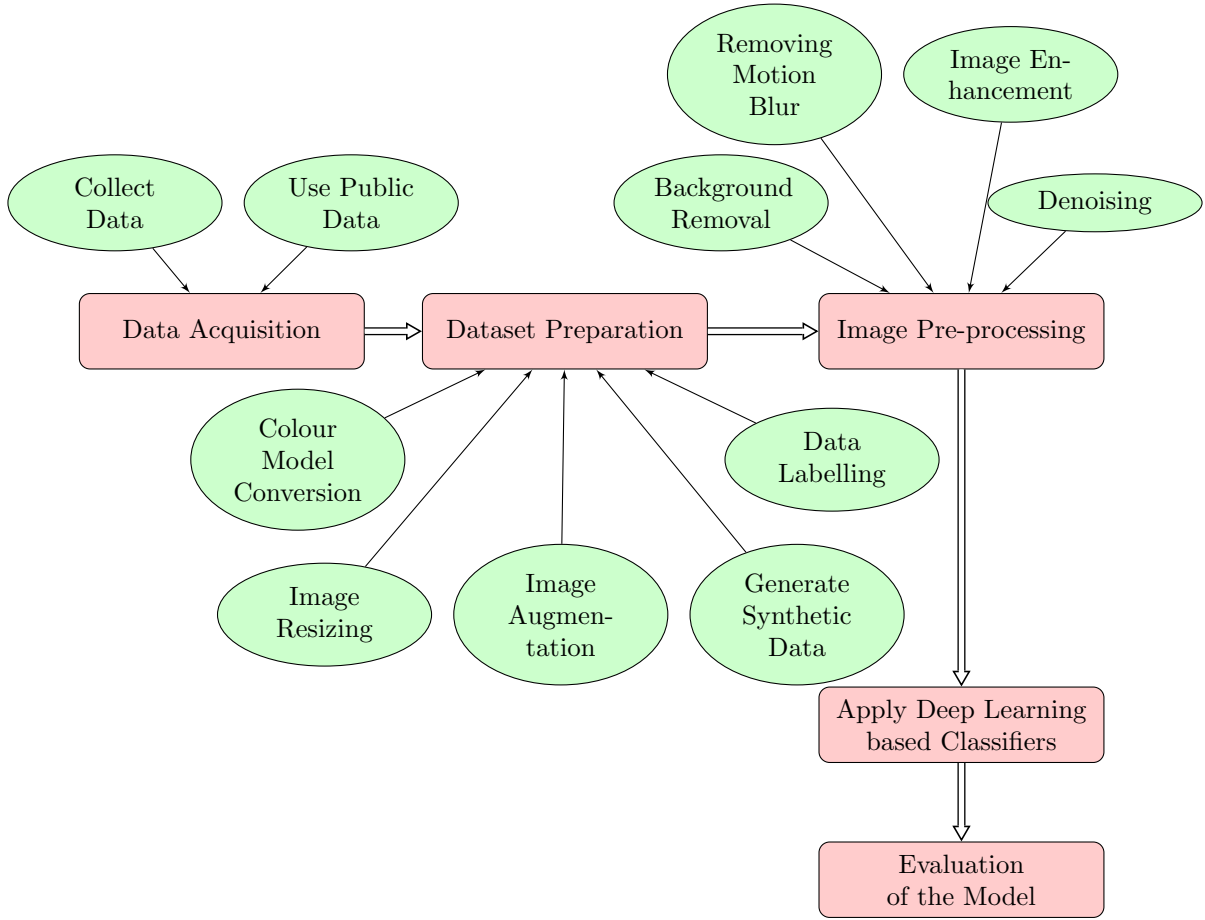
\begin{figure}[t]
    \centering
    \setstretch{1.0}
\begin{tikzpicture}
\node[elli] (Collect_Data){Collect Data};
\node[elli, right of = Collect_Data, xshift = 6em](Public_Data){Use Public Data};
\node[block, below of = Collect_Data, xshift = 4em, yshift =-2em] (Data_Acquisition){Data Acquisition};
\draw [arrow] (Collect_Data) -- (Data_Acquisition);
\draw [arrow] (Public_Data) -- (Data_Acquisition);

\node[block, right of = Data_Acquisition, xshift = 10em] (Dataset_Preparation){Dataset Preparation};

\draw [vecArrow] (Data_Acquisition) -- (Dataset_Preparation);

\node[elli, below of = Dataset_Preparation, xshift = -10em, yshift = -2em](Colour_Model_Conversion){Colour Model Conversion};
\node[elli, below of = Colour_Model_Conversion, xshift = 1em, yshift = -3em] (Resize_Image){Image Resizing};
\node[elli, right of = Resize_Image, xshift = 6em](Image_Augmentation){Image Augmentation};
\node[elli, below of = Dataset_Preparation, xshift = 10em, yshift =-2em](Data_Labelling){Data Labelling};
\node[elli, right of = Image_Augmentation, xshift = 6em] (Generate_Synthetic_Data){Generate Synthetic Data};

\draw [arrow] (Colour_Model_Conversion) -- (Dataset_Preparation);
\draw [arrow] (Resize_Image) -- (Dataset_Preparation);
\draw [arrow] (Image_Augmentation) -- (Dataset_Preparation);
\draw [arrow] (Data_Labelling) -- (Dataset_Preparation);
\draw [arrow] (Generate_Synthetic_Data) -- (Dataset_Preparation);

\node[block, right of = Dataset_Preparation, xshift = 12em] (Image_Pre_processing){Image Pre-processing};

\node[elli, above of = Image_Pre_processing, xshift = -9em, yshift = 2em](Background_Removal){Background Removal};
\node[elli, above of = Background_Removal, xshift = 2em, yshift = 2em] (motion_blur){Removing Motion Blur};
\node[elli, right of = motion_blur, xshift = 6em](Image_enhancement){Image Enhancement};
\node[elli, above of = Image_Pre_processing, xshift = 5em, yshift = 2em](denoising){Denoising};

\draw [arrow] (Background_Removal) -- (Image_Pre_processing);
\draw [arrow] (motion_blur) -- (Image_Pre_processing);
\draw [arrow] (denoising) -- (Image_Pre_processing);
\draw [arrow] (Image_enhancement) -- (Image_Pre_processing);
\draw [vecArrow] (Dataset_Preparation) -- (Image_Pre_processing);

\node[block, below of = Image_Pre_processing, yshift = -12em] (Deep_Learning_Based_Classifier){Apply Deep Learning based Classifiers};
\node[block, below of = Deep_Learning_Based_Classifier, yshift = -3em] (Evaluation_of_Model){Evaluation of the Model};

\draw [vecArrow] (Image_Pre_processing) -- (Deep_Learning_Based_Classifier);
\draw [vecArrow] (Deep_Learning_Based_Classifier) -- (Evaluation_of_Model);

\end{tikzpicture}
\caption[A summary workflow of weed detection techniques using deep learning]{A summary workflow of weed detection techniques using deep learning.}
    \label{fig4:DL_workflow}
\end{figure}

Conventional ML techniques require substantial domain expertise to construct a feature extractor from raw data.  On the other hand, the DL approach uses a representation-learning method where a machine can automatically discover the discriminative features from raw data for classification or object detection problems \parencite{lecun2015deep}.  A machine can learn to classify directly from images, text and sounds \parencite{patterson2017deep}.  The ability to extract the features that best suit the task automatically is also known as feature learning.  As deep learning is a hierarchical architecture of learning, the features of the higher levels of the hierarchy are composed of lower-level features \parencite{najafabadi2015deep,hinton2006fast}.\par

Several popular and high performing network architectures are available in deep learning. Two of the frequently used architectures are Convolutional Neural Networks (CNNs) and Recurrent Neural Networks (RNNs) \parencite{lecun2015deep, hosseini2020deep}.  Although CNNs are used for other types of data, the most widespread use of CNNs is to analyse and classify images.  The word convolution refers to the filtering process.  A stack of convolutional layers is the basis of CNN. Each layer receives the input data, transform, or convolve them and output to the next layer.  This convolutional operation eventually simplifies the data so that it can be better processed and understood.  RNNs have a built-in feedback loop, which allows them to act as a forecasting engine. Feed-forward or CNN take a fixed size input and produces a fixed size output.  The signal flow of the feed-forward network is unidirectional, i.e., from input to output.  They cannot even capture the sequence or time-series information.  RNNs overcome the limitation.  In RNN, the current inputs and outputs of the network are influenced by prior input.  Long Short-Term Memory (LSTM) is a type of RNN \parencite{lecun2015deep}, which has a memory cell to remember important prior information, thus can help improving the performance.  Depending on the network architecture, DL has several components like convolutional layers, pooling layers, activation functions, dense/fully connected layers, encoder/decoder schemes, memory cells, gates etc. \parencite{patterson2017deep}.\par

For image classification, object detection, and localisation, DL algorithms have many advantages over traditional ML approaches. Because of the strong feature learning capabilities, DL methods can effectively extract discriminative features of crops and weeds. Also, with increasing data, the performance of traditional ML approaches has become saturated. Using large dataset, DL techniques show superior performance compared to traditional ML techniques \parencite{alom2019state}. This characteristic is leading to the increasing application of DL approaches. Many of the research reports in Section \ref{Deep_Learning_Architecture} show  comparisons between DL and other ML approaches to detect weeds in crops. Figure \ref{fig4:DL_workflow} gives an overview of DL-based weed detection and recognition techniques.\par

Not all the steps outlined in Figure \ref{fig4:DL_workflow} need to be present in every method. Four major steps are followed in this process. They are Data Acquisition, Dataset Preparation/Image Pre-processing, Classification and Evaluation. In this paper, we describe the steps used in different research work to discriminate between weeds and crops using DL techniques.\par

\section{Paper Selection Criteria in this Survey} \label{paper_selection}

To overview the state of research, we have undertaken a comprehensive literature review. The process involved two major steps: (i) searching and selecting related studies and (ii) detailed analysis of these studies. The main research question is: What is the role of deep learning techniques for detecting, localising and classifying weeds in crops? For collecting the related work based on this research question, we applied a keyword-based search in Google Scholar, Web of Science, IEEE Xplore, Scopus, ScienceDirect, Multidisciplinary Digital Publishing Institute (MDPI), Springer and Murdoch University Library databases for journal articles and conference papers. We have applied a keyword search from 2010 to 30 August 2020. Table \ref{table:search_result} shows the number of search results for the search query.\par

\begin{table}[t]
\centering
\caption{Number of documents resulted for the queries indicated}
\label{table:search_result}
\scriptsize
    \begin{tabularx}{\textwidth}{|M{0.4cm}|P{2.5cm}|p{9cm}|Y|}
    \hline
    No. & Academic Research Databases & \multicolumn{1}{c|}{Search Query} & Number of Retrieved Documents \\ \hline
    1.  & Google   Scholar                                 & {[}\enquote{Weed Detection} OR \enquote{Weed management} OR \enquote{Weed Classification}{]} AND   {[}\enquote{Deep Learning} OR \enquote{Deep Machine Learning} OR \enquote{Deep Neural Network}{]}                       & 998                           \\ \hline
2.  & Web of   Science                                 & (Weed Detection OR Weed   management OR Weed Classification) AND (Deep Learning OR Deep Machine   Learning OR Deep Neural Network)                                             & 124                           \\ \hline
3.  & IEEE   Xplore                                    & (((\enquote{All   Metadata}:\enquote{Deep Learning}) OR \enquote{All   Metadata}:\enquote{Deep Machine Learning}) OR \enquote{All   Metadata}:\enquote{Deep Neural Network}) AND Weed detection                                & 22                            \\ \hline
4.  & ScienceDirect                                    & (\enquote{Weed Detection} OR   \enquote{Weed management} OR \enquote{Weed Classification}) AND   (\enquote{Deep Learning} OR \enquote{Deep Machine Learning} OR \enquote{Deep Neural Network})                               & 87                            \\ \hline
5.  & Scopus                                           & ((Weed AND detection) OR (Weed   AND Management) OR (Weed AND Classification)) AND ((Deep AND Learning) OR   (Deep AND Machine AND Learning) OR (Deep AND Neural AND Network)) & 118                           \\ \hline
6.  & MDPI                                             & (\enquote{Weed Detection} OR   \enquote{Weed management} OR \enquote{Weed Classification}) AND   (\enquote{Deep Learning} OR \enquote{Deep Machine Learning} OR \enquote{Deep Neural Network})                               & 76                            \\ \hline
7.  & SpringerLink                                     & (\enquote{Weed Detection} OR   \enquote{Weed management} OR \enquote{Weed Classification}) AND   (\enquote{Deep Learning} OR \enquote{Deep Machine Learning} OR \enquote{Deep Neural Network})                               & 46                            \\ \hline
8.  & Murdoch   University Library                     & (\enquote{Weed Detection} OR   \enquote{Weed management} OR \enquote{Weed Classification}) AND   (\enquote{Deep Learning} OR \enquote{Deep Machine Learning} OR \enquote{Deep Neural Network})                               & 179                           \\
\hline
    \end{tabularx}
\end{table}

\begin{figure}[t]
    \centering
    \begin{tikzpicture}
    \begin{axis}[
            ybar,
            bar width=.3cm,
            width=.8\textwidth,
            height=.4\textwidth,
            legend style={at={(0.5,1)}, anchor=north,legend columns=-1},
            symbolic x coords={2010,2011,2012,2013,2014,2015,2016,2017,2018,2019,2020},
            xtick=data,
            nodes near coords,
            nodes near coords align={vertical},
            ymin=0,
            ylabel={Number of Publication},
            legend image code/.code={ \draw [#1] (0cm,-0.1cm) rectangle (0.2cm,0.25cm); },
        ]
        \addplot coordinates { (2010, 11) (2011, 3) (2012, 4) (2013, 8) (2014, 12) (2015, 13) (2016, 47) (2017, 91) (2018, 193) (2019, 331) (2020, 275)};
        \addplot coordinates {  (2013, 1) (2014, 1) (2015, 2) (2016, 5) (2017, 16) (2018, 30) (2019, 46) (2020, 38)};
        \legend{Searched document, DL based article}
    \end{axis}
    \end{tikzpicture}
    \caption{The number of selected publications on DL-based weed detection approach from 2010 to 30 August 2020}
    \label{fig:num_of_pub_2010}
\end{figure}
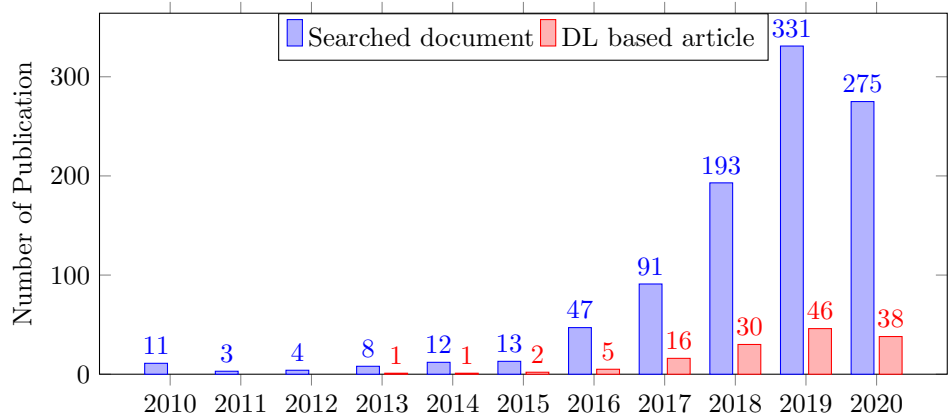

After searching the above databases, duplicated documents were removed: that provided 988  documents. We further identified and counted those using DL-based methodology. In Figure \ref{fig:num_of_pub_2010}, we show the total number of papers which used DL between 2010 to 30 August 2020. This shows that before 2016, the number of publications in this area was very small, but that there is an upward trend in the number of papers from 2016. For this reason, articles published from 2016 and onward were used in this survey.\par

\section{An Overview and Taxonomy of Deep Learning-based Weed Detection Approaches} \label{overview}

\begin{figure}[H]
    \centering
    \includegraphics[width=0.85\textwidth]{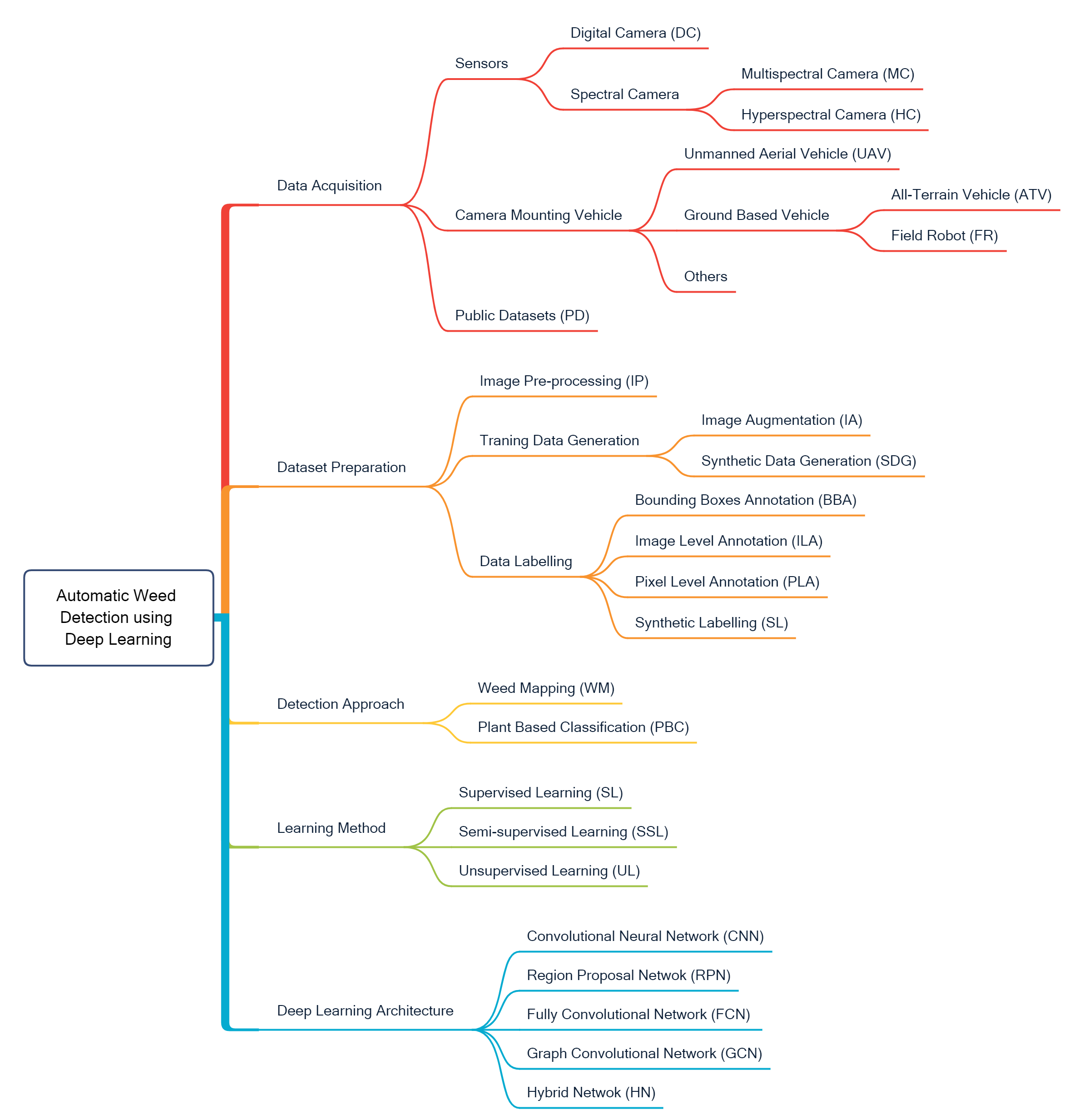}
    \caption[An overall taxonomy of deep learning-based weed detection techniques]{An overall taxonomy of deep learning-based weed detection techniques}
    \label{fig:texonomy_DL_techniques}
\end{figure}

An overall taxonomy of DL-based weed detection techniques is shown in Figure \ref{fig:texonomy_DL_techniques}. The papers covered in this survey are categorised using this taxonomy and listed in Table \ref{tab:different_DL_approach}.\par 

The related publications have been analysed based on the taxonomy in Figure \ref{fig:texonomy_DL_techniques}. Here, the data acquisition process, sensors and mounting vehicles are highlighted. Moreover, an overview of the dataset preparation approaches, i.e., image pre-processing, data generation and annotation are also given. While analysing these publications, it has been found that the related works either generate a weed map for the target site or a classification for each of the plants (crops/weeds). For developing the classifiers, the researchers applied supervised, unsupervised or semi-supervised learning approaches. Depending on the learning approaches and the research goal, different DL architectures were used. An overview of the related research is provided in Table \ref{tab:different_DL_approach}. It shows the crop and weed species selected for experimental work, the steps taken to collect and prepare the datasets, and the DL methods applied in the research.

\begin{landscape}

\begin{center}
\scriptsize
\setstretch{1.0}
\begin{longtable}{|L{0.13\textheight}|L{0.12\textheight}|L{0.33\textheight}|L{0.18\textheight}|L{0.13\textheight}|}
\caption{An overview of different DL approaches used in weed detection}
\label{tab:different_DL_approach}\\
\hline
\multicolumn{1}{|>{\centering\arraybackslash}L{0.13\textheight}|}{\textbf{Reference}} & \multicolumn{1}{>{\centering\arraybackslash}L{0.1\textheight}|}{\textbf{Crop}} & \multicolumn{1}{>{\centering\arraybackslash}L{0.3\textheight}|}{\textbf{Weed Species}} & \multicolumn{1}{>{\centering\arraybackslash}L{0.15\textheight}|}{\textbf{DL Architectures Applied}} & \multicolumn{1}{>{\centering\arraybackslash}L{0.13\textheight}|}{\textbf{Operations Performed (based on Figure \ref{fig:texonomy_DL_techniques})}} 
\\
\hline
\endfirsthead
\multicolumn{5}{c}%
{\tablename\ \thetable\ -- \textit{Continued from previous page}} \\
\hline
\multicolumn{1}{|>{\centering\arraybackslash}L{0.13\textheight}|}{\textbf{Reference}} & \multicolumn{1}{>{\centering\arraybackslash}L{0.1\textheight}|}{\textbf{Crop}} & \multicolumn{1}{>{\centering\arraybackslash}L{0.3\textheight}|}{\textbf{Weed Species}} & \multicolumn{1}{>{\centering\arraybackslash}L{0.15\textheight}|}{\textbf{DL Architectures Applied}} & \multicolumn{1}{>{\centering\arraybackslash}L{0.13\textheight}|}{\textbf{Operations Performed (based on Figure \ref{fig:texonomy_DL_techniques})}} \\
\hline
\endhead
\hline \multicolumn{5}{r}{\textit{Continued on next page}} \\
\endfoot
\endlastfoot
\textcite{espejo2020towards} & Tomato, Cotton & Black nightshade, velvetleaf & Modified Xception, Inception-ResNet, VGGNet, MobileNet, DenseNet  & DC; (IP, IA, ILA); PBC\\ \hline

\textcite{wang2020semantic} & Sugar beet, Oilseed  & Not specified  & FCN & (DC, FR); (IP, IA, ILA); PBC  \\ \hline
\textcite{le2020performances}  & Canola, corn, radish & Not specified & Filtered Local Binary Pattern with Contour Mask and Coefficient k (k-FLBPCM), VGG-16, VGG-19,   ResNet-50, Inception-v3 & (ATV, MC); (IP, IA, ILA); PBC \\ \hline

\textcite{hu2020graph}  & Not specified & Chinee apple, Lantana, Parkinsonia, Parthenium, Prickly acacia, Rubber vine, Siam weed, Snake weed & Inception-v3, ResNet-50, DenseNet-202, Inception-ResNet-v2, GCN & PD; IP; PBC \\ \hline

\textcite{umamaheswari2020encoder} & Carrot & Not specified & SegNet-512, SegNet-256 & PD; IA; PBC \\ \hline

\textcite{huang2020deep} & Rice & \textit{Leptochloa chinensis}, Cyperus iria, Digitaria sanguinalis (L). Scop, Barnyard Grass & FCN & (DC, UAV); (IP, PLA); WM \\ \hline

\textcite{gao2020deep} & Sugar beet & Convolvulus sepium (hedge bindweed) & YOLO-v3, tiny YOLO-v3 & DC; (IA, BBA); PBC \\ \hline
\textcite{veeranampalayam2020comparison} & Soybean & Waterhemp, Palmer amaranthus, common lambsquarters, velvetleaf, foxtail species & Single-Shot Detector (SSD), Faster R-CNN & (DC, UAV); (IP, IA, BBA); WM \\ \hline

\textcite{jiang2020cnn} & Corn, lettuce, radish & Cirsium setosum, Chenopodium album, bluegrass, sedge, other unspecified weed & GCN &  PD; (IP, ILA); PBC \\ \hline

\textcite{bosilj2020transfer} & Sugar Beets, Carrots, Onions & Not specified & SegNet &  PD; PLA; PBC \\ \hline

\textcite{yan2020classification} & Paddy & Alternanthera philoxeroides, Eclipta prostrata, Ludwigia adscendens, Sagittaria trifolia, Echinochloa crus-galli, Leptochloa chinensis & AlexNet & DC; ILA; PBC \\ \hline

\textcite{zhang2020weed} & Wheat & Cirsium Setosum, Descurainia Sophia, Euphorbia Helioscopia, Veronica Didyma, Avena Fatu & YOLO-v3, Tiny YOLO-v3 & (DC, UAV); (IP, PLA); PBC \\ \hline

\textcite{lottes2020robust} & Sugar beet & Dicot weeds, grass weeds & FCN & MC; (IP, PLA); PBC \\ \hline

\textcite{trong2020late} & Not specifies & 12 species of \enquote{Plant Seedlings dataset}, 21 species of \enquote{CNU weeds dataset} & NASNet, ResNet, Inception–ResNet, MobileNet, VGGNet& DC; ILA, PD \\ \hline

\textcite{patidar2020weed} & Not specified & Scentless Mayweed, Chickweed, Cranesbill, Shepherd's Purse, Cleavers, Charlock, Fat Hen, Maise, Sugar beet, Common wheat, Black-grass, Loose Silky-bent & Mask R-CNN & PD; PLA; PBC \\ \hline 
\textcite{ramirez2020deep} & Sugar beet & Not specified & DeepLab-v3, SegNet, U-Net & (MC, UAV); (IP, PLA)\\ \hline

\textcite{osorio2020deep} & Lettuce & Not specified & YOLO-v3, Mask R-CNN, SVM & (MC, UAV); (IP, PLA) \\ \hline

\textcite{lam2020open} & Grasslands & Rumex obtusifolius & VGG-16 & (DC, UAV); (IP, PLA) \\ \hline

\textcite{sharpe2020goosegrass} & Strawberry, Tomato & Goosegrass & Tiny YOLO-v3 & DC; (IP, BBA) \\ \hline

\textcite{petrich2019detection} & Not specified & Colchicum autumnale & U-Net & (DC, UAV); (IP, IA, BBA) \\ \hline

\textcite{czymmek2019vision} & Carrot & Not specified & Faster YOLO-v3, tiny YOLO-v3 & (DC, FR); ILA; PBC \\ \hline

\textcite{partel2019smart} & Blueberry & Not specified & Faster R-CNN, YOLO-v3, ResNet-50,   ResNet-101, Darknet-53 & (DC, ATV); (IP, ILA); PBC \\ \hline

\textcite{partel2019development} & Pepper & Portulaca weeds & Tiny YOLO-v3, YOLO-v3 & (DC, ATV); (IP, BBA); PBC \\ \hline

\textcite{olsen2019deepweeds} & Not specified & Chinee apple, Lantana, Parkinsonia, Parthenium, Prickly acacia, Rubber vine, Siam weed, Snake weed & Inception-v3, ResNet-50 & (DC, FR); (IP, ILA); PBC \\ \hline

\textcite{kounalakis2019deep} & Clover, grass & Broad-leaved dock & AlexNet, VGG-F, VGG-VD-16,   Inception-v1, ResNet-50, ResNet-101 & (DC, FR); PLA; PBC \\ \hline

\textcite{rasti_ahmad_samiei_belin_rousseau_2019} & Mache salad & Not specified & Scatter Transform, Local Binary Pattern (LBP), GLCM,   Gabor filter, CNN & (DC, FR); (IP, SDG, BBA); PBC \\ \hline

\textcite{sarvini2019performance} & Chrysanthemum & Para grass, Nutsedge & SVM, Artificial Neural Network (ANN), CNN & DC; (IP, IA, ILA); PBC \\ \hline

\textcite{ma2019fully} & Rice & Sagittaria trifolia & SegNet, FCN, U-Net & DC; (IP, BBA); PBC \\ \hline

\textcite{asad_bais_2019} & Canola & Not specified & U-Net, SegNet & (DC, ATV); (IP, IA, PLA); PBC \\ \hline

\textcite{yu2019deep} & Bermudagrass & Hydrocotyle spp., Hedyotis cormybosa, Richardia scabra & VGGNet, GoogLeNet, DetectNet & DC; (IP, ILA); PBC \\ \hline

\textcite{abdalla2019fine} & Oilseed & Not specified & FCN & DC; (IA, PLA); WM \\ \hline

\textcite{yu2019weed} & Perennial ryegrass & dandelion, ground ivy, spotted spurge & AlexNet, VGGNet, GoogLeNet, DetectNet &  DC; (IP, ILA); PBC \\ \hline

\textcite{liang2019low} & Not specified & Not specified & CNN, Histogram of oriented Gradients (HoG), LBP & (DC, UAV); (IP, ILA); PBC \\ \hline

\textcite{sharpe2019detection} & Strawberry & Carolina geranium & VGGNet, GoogLeNet, DetectNet & DC; (IP, BBA); PBC \\ \hline

\textcite{fawakherji2019crop} & Sunflower, carrots, sugar beets & Not specified & SegNet, U-Net, BonNet, FCN8 & (DC, FR, PD); PLA; PBC \\ \hline

\textcite{valente2019detecting} & Grassland & Rumex obtusifolius & AlexNet & (DC, UAV); (IP, BBA); PBC \\ \hline

\textcite{chechlinski2019system} & Beet, cauliflower, cabbage, strawberry & Not specified & Hybrid Network & (DC, ATV); (IP, IA, PLA); PBC\\ \hline

\textcite{brilhador2019classification} & Carrot & Not specified & U-Net & PD; (IA, PLA); PBC \\ \hline

\textcite{binguitchacrops} & Maise, common wheat, sugar beet & Scentless Mayweed, common chickweed, shepherd's purse, cleavers, Redshank, charlock, fat hen, small-flowered Cranesbill, field pansy, black-grass, loose silky-bent & ResNet-101 &  PD, (IP, IA, BBA); PBC \\ \hline

\textcite{jiang2019deepseedling} & Cotton & Not specified & Faster R-CNN & DC, (IP, BBA); PBC \\ \hline

\textcite{adhikari2019learning} & Paddy & Wild millet & ESNet,   U-Net, FCN-8s, and DeepLab-v3, Faster R-CNN, EDNet & DC; (IP, IA, PLA); PBC \\ \hline

\textcite{farooq2019multi} & Sugar beet & Alli, hyme, hyac, azol, other unspecified weeds & CNN, FCN, LBP, superpixel based  LBP, FCN-SPLBP & HC; (IP, PLA); PBC \\ \hline

\textcite{knoll2019real} & Carrot & Not specified &  CNN & DC; (IP, PLA); PBC \\ \hline

\textcite{dos2019unsupervised} & Soybean & grass, broadleaf weeds, Chinee apple, Lantana, Parkinsonia, Parthenium, Prickly acacia, Rubber vine, Siam weed, Snake weed & Joint Unsupervised LEarning (JULE), DeepCluster & PD; PBC \\ \hline

\textcite{rist2019weed} & Not specified & Gamba grass & U-Net & SI; (IP, PLA)\\ \hline

\textcite{skovsen2019grassclover} & Clover & Grass & FCN-8s &  DC, (IP, PLA); PBC\\ \hline

\textcite{zhang2018broad} & Pasture & Not specified & CNN, SVM & (DC, ATV); (IP, IA, ILA); PBC \\ \hline

\textcite{kounalakis2018robotic} & Grasslands & Broad-leaved dock & AlexNet, VGG-F, GoogLeNet & (DC, FR); BBA; PBC \\ \hline

\textcite{huang2018semantic} & Rice & Not specified & FCN, SVM & (DC, UAV); BBA; WM \\ \hline

\textcite{teimouri2018weed} & Not specified & Common field speedwell, field pansy, common chickweed, fat-hen, fine grasses (annual meadow-grass, loose silky-bent), blackgrass, hemp-nettle, shepherd’s purse, common fumitory, scentless mayweed, cereal, brassicaceae, maise, polygonum, oat (volunteers), cranesbill, dead-nettle, common poppy & Inception-v3 & DC; (IP, ILA); PBC \\ \hline

\textcite{umamaheswari2018weed} & Carrot & Not specified & GoogleNet &  PD, (IA, BBA); PBC \\ \hline

\textcite{suh2018transfer} & Sugar beets & Volunteer potato & AlexNet, VGG-19, GoogLeNet,   ResNet-50, ResNet-101, Inception-v3 & (DC, ATV); (IP, IA, ILA); PBC\\ \hline

\textcite{farooq2018analysis} & Not specified & Hyme, Alli, Azol, Hyac & CNN & HC, (IP, IA, BBA); PBC\\ \hline

\textcite{bah2018deep} & Spinach, bean & Not specified & ResNet-18 & (DC, UAV); (IP, IA, BBA); WM \\ \hline

\textcite{farooq2018weed} & Not specified & Hyme, Alli, Azol, Hyac & CNN, HoG & HC, (IP, ILA); PBC\\ \hline

\textcite{lottes2018fully} & Sugar beet & Not specified & FCN & (MC, FR); (IP, PLA); PBC \\ \hline

\textcite{sa2018weedmap} & Sugar beet & Galinsoga spec., Amaranthus retroflexus, Atriplex spec., Polygonum spec., Gramineae (Echinochloa crus-galli, agropyron, others.), Convolvulus arvensis, Stellaria media, Taraxacum spec. & SegNet & (MC, UAV); PLA; WM \\ \hline

\textcite{huang2018fully} & Rice & Not specified & CNN, FCN & (DC, UAV); (IP, PLA); WM \\ \hline

\textcite{huang2018accurate} & Rice & Not specified & FCN-8s, FCN-4s, DeepLab & (DC, UAV) (IP, PLA); WM \\ \hline

\textcite{chavan2018agroavnet} & Maise, common wheat, sugar beet & Scentless Mayweed, common chickweed, shepherd's purse, cleavers, Redshank, charlock, fat hen, small-flowered Cranesbill, field pansy, black-grass, loose silky-bent & AlexNet, VGGNet, Hybrid Network &  PD; PBC \\ \hline

\textcite{nkemelu2018deep} & Maise, common wheat, sugar beet & Scentless Mayweed, common chickweed, shepherd's purse, cleavers, Redshank, charlock, fat hen, small-flowered Cranesbill, field pansy, black-grass, loose silky-bent & KNN, SVM, CNN &  PD; (IP, BBA); PBC \\ \hline

\textcite{sa2017weednet} & Sugar beet & Not specified & SegNet & (MC, UAV), (IP, BBA), WM \\ \hline

\textcite{andrea2017precise} & Maise & Not specified & LeNET, AlexNet, cNET, sNET & DC; (IP, IA, PLA); PBC\\ \hline

\textcite{dyrmann2017roboweedsupport} & Winter wheat & Not specified & FCN & (DC, ATV); (IP, BBA); PBC\\ \hline

\textcite{dos2017weed}  & Soybean & Grass, broadleaf weeds & AlexNet, SVM, Adaboost – C4.5, Random Forest & (DC, UAV); (IP, ILA);  PBC\\ \hline

\textcite{tang2017weed} & Soybean  & Cephalanoplos, digitaria, bindweed &  Back propagation neural network, SVM, CNN & DC; (IP, ILA); PBC\\ \hline

\textcite{milioto2017real} & Sugar beet & Not specified & CNN &  (DC, UAV); (IP, PLA); PBC \\ \hline

\textcite{pearlstein2016convolutional} & Lawn grass & Not specified & CNN & (DC, FR); (IP, SDG, BBA); PBC   \\ \hline

\textcite{di2017automatic} & Sugar beet & Capsella bursa-pastoris, galium aparine & SegNet & SDG, PBC \\ \hline

\textcite{dyrmann2016plant} & Tobacco, thale cress, cleavers, common Poppy, cornflower, wheat, maise, sugar beet, cabbage, barley & Sherpherd's-Purse , chamomile, knotweed family, cranesbill, chickweed, veronica, fat-hen, narrow-leaved grasses, field pancy, broad-leaved grasses, annual nettle, black nightshade & CNN & PD; (IP, IA); PBC\\ \hline


\end{longtable}
\end{center}
\vspace{-4.5em}
\begin{center}
\end{center}
\end{landscape}

\section{Data Acquisition} \label{Data_Acquisition}

DL based weed detection and classification techniques require an adequate amount of labelled data. Different modalities of data are collected using various types of sensors that are mounted on a variety of platforms. Below we discuss the popular ways of weed data collection.\par

\subsection{Sensors and Camera Mounting Vehicle}

\subsubsection{Unmanned Aerial Vehicles (UAVs)}

Unmanned Aerial Vehicles are often  used for data acquisition in agricultural research. Generally, UAVs are used for mapping weed density across a field by collecting RGB images \parencite{huang2020deep,huang2018fully,huang2018semantic, petrich2019detection} or multispectral images \parencite{sa2018weedmap, sa2017weednet, patidar2020weed, osorio2020deep, ramirez2020deep}. In addition, UAVs can be used to identify crop rows and map weeds within crop rows by collecting RGB (Red, Green and Blue color) images \parencite{bah2018deep}. \textcite{valente2019detecting} used a small quad-rotor UAV for recording images from grassland to detect broad-leaved dock (\textit{Rumex obtusifolius}). As UAVs fly over the field at a certain height, the images captured by them cover a large area. Some of the studies split the images into smaller patches and use the patches to distinguish between weeds and crop plants \parencite{milioto2017real,dos2017weed, veeranampalayam2020comparison}. However, the flight altitude can be maintained at a low height, e.g. 2 meters, so that each plant can be labelled as either a weed or crop \parencite{zhang2020weed, osorio2020deep}. \textcite{liang2019low} collected image data using a drone by maintaining an altitude of 2.5 meters. \textcite{huang2018accurate} collected images with a resolution of 3000$\times$4000 pixels using a sequence of forward-overlaps and side-overlaps to cover the entire field. \textcite{lam2020open} flew DJI Phantom 3 and 4 Pro drones with a RGB camera at three different heights (10, 15 and 20 m) to determine the optimal height for weed detection.\par

\subsubsection{Field Robots (FRs)}
Various types of field robot can also be used to collect images. A robotic vehicle can carry one or more cameras. As previously discussed, robotic vehicles are used to collect RGB images by mounted digital cameras \parencite{czymmek2019vision,olsen2019deepweeds,kounalakis2019deep,rasti_ahmad_samiei_belin_rousseau_2019,fawakherji2019crop}. Mobile phone in-built cameras have also been used for such data collection. For example, an iPhone 6 was used to collect video data by mounting it on a Robotic Rover \parencite{pearlstein2016convolutional}. A robotic platform called “BoniRob” has been used to collect multi-spectral images from the field \parencite{lottes2018fully, lottes2020robust}. \textcite{kounalakis2018robotic} used three monochrome cameras mounted on a robot to take images. They argued that, in most cases, weeds are green, and so are the crops. There is no need to use colour features to distinguish them.\par

\subsubsection{All-Terrain Vehicles (ATVs)}

To collect images from the field, all-terrain vehicles have also been used. ATVs can be mounted with different types of camera \parencite{dyrmann2017roboweedsupport, zhang2018broad, asad_bais_2019, partel2019smart, chechlinski2019system}. \textcite{le2020performances} used a combination of multi-spectral and spatial sensors to capture data. Even multiple low-resolution webcams have been used on an ATV \parencite{partel2019development}. To maintain specific height with external lighting conditions, and illumination, custom made mobile platforms have been used to carry the cameras for capturing RGB images \parencite{suh2018transfer,skovsen2019grassclover}. When it is not possible to use any vehicle to collect images at a certain height, tripods can be used as an alternative \parencite{abdalla2019fine}.\par

\subsubsection{Collect Data without Camera Mounting Devices}

On a few occasions, weed data have been collected by cameras without being mounted on a vehicle. As such, video data are collected using handheld cameras \parencite{yu2019deep, sarvini2019performance, ma2019fully, yu2019weed, teimouri2018weed, tang2017weed, espejo2020towards,gao2020deep, adhikari2019learning, knoll2019real, yan2020classification, jiang2019deepseedling,sharpe2020goosegrass}. \textcite{sharpe2019detection} collected their data by maintaining a certain height (130 cm) from the soil surface.
Brimrose VA210 filter and JAI BM-141 cameras have been used to collect hyperspectral images of weeds and crops without using any platform \parencite{farooq2018analysis, farooq2019multi, farooq2018weed}. \textcite{andrea2017precise} manually focused a camera on the target plants in such a way that it could capture images, including all the features of these plants. In \textcite{trong2020late}, they focus the camera on many parts of weeds, such as flowers, leaf, fruits, or the full weeds structure.

\subsection{Satellite Imagery}

\textcite{rist2019weed} use the Pleiades-HR 1A to collect high-resolution 4-band (RGB+NIR) imagery over the area of interest. They made use of high-resolution satellite images and applied masking to indicate the presence of weeds.

\subsection{Public Datasets}

There are several publicly available crop and weed datasets that can be used to train the DL models. \textcite{chebrolu2017agricultural} developed a dataset containing weeds in sugar beet crops. Another annotated dataset containing images of crops and weeds collected from fields has been made available by \textcite{haug2014crop}. A dataset of annotated (7853 annotations) crops and weed images was developed by \textcite{sudars2020dataset}, which comprises 1118 images of six food crops and eight weed species. \textcite{leminen2020open} developed a dataset containing 7,590 RGB images with 315,038 plant objects, representing 64,292 individual plants from 47 different species. These data were collected in Denmark and made available for further use. A summary of the publicly available datasets related to weed detection and plant classification is listed in Table \ref{tab:public_datasets}.\par

We have listed nineteen datasets in Table \ref{tab:public_datasets} which are available in this area, and can be used by researchers. Amongst these datasets, researchers will need to send a request to the owner of \enquote{Perennial ryegrass and weed}, \enquote{CNU Weed Dataset} and \enquote{Sugar beet and hedge bindweed} dataset to obtain the data. Other datasets can be downloaded directly on-line. Most of the datasets contain RGB images of food crops and weeds from different parts of the world. The RGB data have generally been collected using high-resolution digital cameras. However, \textcite{teimouri2018weed} used a point grey industrial camera. While acquiring data for the \enquote{DeepWeeds} dataset, the researchers added a  \enquote{Fujinon CF25HA-1} lens with their \enquote{FLIR Blackfly 23S6C} camera and mounted the camera on a weed control robot (\enquote{AutoWeed}). \textcite{chebrolu2017agricultural} and \textcite{haug15} employed \enquote{Bonirob} (an autonomous field robot) to mount the multi-spectral cameras. \enquote{Carrots 2017} and \enquote{Onions 2017} datasets were also acquired using a multi-spectral camera, namely the \enquote{Teledyne DALSA Genie Nano}. These researchers used a manually pulled cart to carry the camera. The \enquote{CNU Weed Dataset} has 208,477 images of weeds collect from farms and fields in the Republic of Korea, which is the highest number among the datasets. Though this dataset exhibits a class imbalance, it contains twenty-one species of weeds from five families. \textcite{skovsen2019grassclover} developed a dataset of red clover, white clover and other associated weeds. The dataset contains 31,600 unlabelled data together with 8000 synthetic data. Their goal was to generate labels for the data using unsupervised or self-supervised approaches. All the other datasets were manually labelled using image level, pixel-wise or bounding box annotation techniques.\par

\begin{sidewaystable}[]
    \centering
        \scriptsize
        \centering
        \caption{List of publicly available crop and weed datasets}
        \label{tab:public_datasets}
        \begin{tabular}{|L{0.14\textwidth}|L{0.07\textwidth}|L{0.11\textwidth}|L{0.04\textwidth}|L{0.06\textwidth}|L{0.08\textwidth}|L{0.04\textwidth}|L{0.05\textwidth}|L{0.04\textwidth}|L{0.16\textwidth}|}
        \hline
        \textbf{Dataset and Reference} & \textbf{Type/Number of Crop} & \textbf{Type/Number of Weed Species} & \textbf{Data Type} & \textbf{Sensor and Mounting Vehicle} & \textbf{Number of Images} & \textbf{Data Annotation} & \textbf{Data Location} & \textbf{Class imbalance?} & \textbf{Source}\\ \hline
        
        Crop/Weed Field Image Dataset \parencite{haug15} & Carrot & Not specified & Multi-spectral image & MC and FR & 60 & PLA & Germany & Yes & \url{https://github.com/cwfid/dataset}\\ \hline

        Dataset of food crops and weed \parencite{sudars2020dataset} & Six crop & Eight weed species & RGB & DC & 1118 & BBA & Latvia & Yes & \url{https://www.ncbi.nlm.nih.gov/pmc/articles/PMC7305380/}\\ \hline
        
        DeepWeeds \parencite{olsen2019deepweeds} & Not specified & Eight weed species & RGB & DC and FR  & 17,509 & ILA & Australia & No & \url{https://github.com/AlexOlsen/DeepWeeds}\\ \hline

        Early crop weed dataset \parencite{espejo2020towards} & Tomato, cotton & Black nightshad, velvetleaf & RGB & 508 & DC & ILA & Greece & Yes & \url{https://github.com/AUAgroup/early-crop-weed}\\ \hline

        Perennial ryegrass and weed \parencite{yu2019weed} & Perennial ryegrass & dandelion, ground ivy, and spotted spurge & RGB & DC & 33086 & ILA & USA & No & \url{https://www.frontiersin.org/articles/10.3389/fpls.2019.01422/full}\\ \hline

        Soybean and weed dataset \parencite{dos2017weed} & Soybean & Grass and broadleaf weeds & RGB & DC and UAV & 400 & ILA & Brazil & Yes & \url{https://www.kaggle.com/fpeccia/weed-detection-in-soybean-crops}\\ \hline

        Open Plant Phenotype Database \parencite{leminen2020open} & Not specified & 46 most common monocotyledon (grass) and dicotyledon (broadleaved) weeds & RGB & DC & 7,590 & BBA & Denmark & No & \url{https://gitlab.au.dk/AUENG-Vision/OPPD} \\ \hline

        Sugar beet and hedge bindweed dataset \parencite{gao2020deep} & Sugar beet & Convolvulus sepium (hedge bindweed) & RGB & DC & 652 & BBA & Belgium & Yes & \url{https://plantmethods.biomedcentral.com/articles/10.1186/s13007-020-00570-z}\\ \hline

        Sugar beet fields dataset \parencite{chebrolu2017agricultural} & Sugar beet & Not specified & Multi-spectral image & MC and  FR & 12340 & PLA & Germany & No & \url{https://www.ipb.uni-bonn.de/2018/10/}\\ \hline

        UAV Sugarbeets 2015-16 Datasets \parencite{chebrolu2018robust} & Sugarbeets & Not specified & RGB & DC and UAV & 675 & PLA & Switzerland &  No & \url{https://www.ipb.uni-bonn.de/data/uav-sugarbeets-2015-16/}\\ \hline

        Corn, lettuce and weed dataset \parencite{jiang2020cnn} & Corn and lettuce & Cirsium setosum, Chenopodium album, bluegrass and sedge & RGB & DC & 6800 & ILA & China & No & \url{https://github.com/zhangchuanyin/weed-datasets}\\ \hline

        Carrot-Weed dataset \parencite{lameski2017weed} & Carrot & Not specified & RGB & DC & 39 & PLA & Republic of Macedonia & Yes & \url{https://github.com/lameski/rgbweeddetection}\\ \hline

        Bccr-segset dataset \parencite{le2020novel} & Canola, corn, radish & Not specified & RGB & DC & 30,000 & ILA & Australia & No & \url{https://academic.oup.com/gigascience/article/9/3/giaa017/5780256\#200419497}\\ \hline

        Carrots 2017 dataset \parencite{bosilj2020transfer} & Carrots & Not specified & Multi-spectral image & MC and manually pulled cart & 20 & PLA & UK & No & \url{https://lcas.lincoln.ac.uk/nextcloud/index.php/s/RYni5xngnEZEFkR}\\ \hline

        Onions 2017 dataset \parencite{bosilj2020transfer} & Onions & Not specified & Multi-spectral image & MC and manually pulled cart & 20 & PLA & UK & No & \url{https://lcas.lincoln.ac.uk/nextcloud/index.php/s/e8uiyrogObAPtcN}\\ \hline

        GrassClover image dataset \parencite{skovsen2019grassclover} & Red clover and white clover & Not specified & RGB &  DC and manually operated platform & 31,600 real and 8000 synthetic images & PLA & Denmark & Yes & \url{https://vision.eng.au.dk/grass-clover-dataset/}\\ \hline

        Leaf counting dataset \parencite{teimouri2018weed} & Not specified & Eighteen weed species & RGB & DC & 9372 & ILA & Denmark & Yes & \url{https://vision.eng.au.dk/leaf-counting-dataset/}\\ \hline

        CNU Weed Dataset \parencite{trong2020late} & Not specified & Twenty one species of weed & RGB & DC & 208,477 & ILA & Republic of Korea & Yes & \url{https://www.sciencedirect.com/science/article/pii/S0168169919319799\#s0025} \\ \hline 

        Plant Seedlings Dataset \parencite{giselsson2017public} &  Three crop & Nine weed species & RGB & DC & 5539 & ILA & Denmark & Yes &  \url{https://www.kaggle.com/vbookshelf/v2-plant-seedlings-dataset}\\ \hline
        
    \end{tabular}
\end{sidewaystable}

\textcite{dyrmann2016plant} use six publicly available datasets containing 22 different plant species to classify using deep learning methods. Several studies proposed an encoder-decoder architecture to distinguish crops and weeds using the Crop Weed Field Image Dataset \parencite{umamaheswari2020encoder,brilhador2019classification,umamaheswari2018weed}. The DeepWeeds dataset \parencite{olsen2019deepweeds} was used by \textcite{hu2020graph} to evaluate their proposed method. In the study of \textcite{jiang2020cnn}, the \enquote{Carrot-Weed dataset} \parencite{lameski2017weed} was used with their own dataset the \enquote{Corn, lettuce and weed dataset}. \textcite{fawakherji2019crop} collected data from a sunflower farm in Italy. To demonstrate the proposed method's generalising ability, they also used two publicly available datasets containing images of carrots, sugar beets and associated weeds. \textcite{bosilj2020transfer} also used those datasets along with the Carrot 2017 and Onion 2017 datasets. The \enquote{Plant Seedlings} dataset is a publicly available dataset containing 12 different plant species. Several studies used this dataset to develop a crop-weed classification model \parencite{chavan2018agroavnet,nkemelu2018deep,binguitchacrops, patidar2020weed}. \textcite{dos2019unsupervised} used DeepWeeds \parencite{olsen2019deepweeds} and \enquote{Soybean and weed} datasets, which are publicly available.

While several datasets are publicly available, they are somewhat site/crop-specific. As such there is no so-called benchmark weed dataset like ImageNet \parencite{deng2009imagenet} and MS COCO \parencite{lin2014microsoft} in this research field, that is widely used in the evaluation.

\section{Dataset Preparation} \label{dataset_preparation}

After acquiring data from different sources, it is necessary to prepare data for training, testing, and to validate models. Raw data is not always suitable for the DL model. The dataset preparation approaches include applying different image processing techniques, data labelling, using image augmentation techniques to increase the number of input data, or impose variations in the data and generating synthetic data for training. Commonly used image processing techniques are - background removal, resizing the collected image, green component segmentation, removing motion blur, de-noising, image enhancement, extraction of colour vegetation indices, and changing the colour model. \textcite{pearlstein2016convolutional} decoded video into a sequence of RGB images and then converted them into grayscale images. In further research, the camera was set to auto-capture mode to collect images in the TIFF format and then these were converted into the RGB colour model \parencite{suh2018transfer}. Using three webcams on an ATV, \textcite{partel2019development} took videos and then converted them into different frames of images. In some occasions, it was necessary to change the image format to accurately train the model, especially when using public datasets. For instance, \textcite{binguitchacrops} converted the \enquote{Plant Seedlings Dataset} \parencite{giselsson2017public} from PNG to JPEG format, as a number of studies have show that the JPEG format is better for training Residual Networks architectures \parencite{ehrlich2019deep}.\par

\subsection{Image Pre-processing}

The majority of relevant studies undertook some level of image processing before providing the data as an input to the DL model. It helps the DL architecture to extract features more accurately. Here we discuss image pre-processing operations used in the related studies.\par

\paragraph{Image Resizing}

\textcite{farooq2018analysis} investigate the performance of Deep Convolutional Neural Networks based on spatial resolution. They used three different special resolutions 30$\times$30, 45$\times$45, and 60$\times$60 pixels. The lower patch size achieved good accuracy and required less time to train the model. To make the processing faster and reduce the computational complexity, most of the studies performed image resizing operations on the dataset before inputting into the DL model. After collecting images from the field, the resolution of the images is reduced based on the DL network requirement. \textcite{yu2019deep} used 1280$\times$720 pixel-sized images to train DetectNet \parencite{tao2016detectnet} architecture and 640$\times$360 pixels for GoogLeNet \parencite{szegedy2015going} and VGGNet \parencite{simonyan2014very} neural networks. The commonly used image sizes (in pixel) are- 64$\times$64  \parencite{bah2018deep,milioto2017real, zhang2018broad,andrea2017precise}, 128$\times$128 \parencite{espejo2020towards,dyrmann2016plant,binguitchacrops}, 224$\times$224 \parencite{olsen2019deepweeds,jiang2020cnn,binguitchacrops}, 227$\times$227 \parencite{valente2019detecting,suh2018transfer}, 228$\times$228 \parencite{le2020performances}, 256$\times$256 \parencite{dos2017weed,tang2017weed,pearlstein2016convolutional,hu2020graph, petrich2019detection}, 320$\times$240 \parencite{chechlinski2019system}, 288$\times$288 \parencite{adhikari2019learning}, 360$\times$360 \parencite{binguitchacrops}.\par 

Images with high resolution are sometimes split into a number of patches to reduce the computational complexity. For instance, in the work of \textcite{rasti_ahmad_samiei_belin_rousseau_2019}, the images were split with a resolution of 5120$\times$3840 into 56 patches. Similar operations were performed by \textcite{huang2018fully,asad_bais_2019,ma2019fully} where they divided the original images into tiles of size 912$\times$1024, 1440$\times$960 and 1000$\times$1000 pixels. \textcite{ramirez2020deep} captured only five images at high resolution using a drone which was then split into small patches of size 480$\times$360 without overlapping and 512$\times$512 with 30\% overlap. \textcite{partel2019smart} collected images using three cameras simultaneously of resolution 640$\times$480 pixels. They then merged those into a single image of 1920$\times$480 pixels which was resized to 1024$\times$256 pixels. \textcite{yu2019weed} scaled down the images of their dataset to 1224$\times$1024 pixels, so that the training did not run low on memory. \textcite{huang2018accurate} used orthomosaic imagery, which is usually quite large. They split the images into small patches of 1000$\times$1000 pixels. In the study of \textcite{sharpe2019detection}, the images were resized to 1280$\times$720 pixels and then cropped into four sub-images. \textcite{osorio2020deep} used 1280$\times$960 pixel size image with four spectral bands.  By applying union operation on the red, green, and near infrared bands, they generated a false green image in order to highlight the vegetation. \textcite{sharpe2020goosegrass} resized the collected image to 1280$\times$853 pixels and then cropped it to 1280$\times$720 pixels. \par

\paragraph{Background Removal}

\textcite{huang2020deep} collected images using a UAV and applied image mosaicing to generate an orthophoto. \textcite{bah2018deep} applied Hough-transform to highlight the aligned pixels and used Otsu-adaptive-thresholding method to differentiate the background and green crops or weeds. On the other hand, for removing the background soil image, \textcite{milioto2017real} applied the Normalised Difference Vegetation Index (NDVI). They also used morphological opening and closing operations to remove the noise and fill tiny gaps among vegetation pixels. To annotate the images manually into respective classes, \textcite{dos2017weed} applied the Simple Linear Iterative Clustering algorithm. This algorithm helps to segment weeds, crops, and background from images. Image pre-processing techniques were also involved in \parencite{sa2017weednet} for having a bounding box around crop plants or weeds and removed the background. They first used image correlation and cropping for alignment and then applied Gaussian blur, followed by a sharpening operation to remove shadows, small debris, etc. Finally, for executing the blob detection process on connected pixels, Otsu’s method was employed. \textcite{lottes2020robust} applied the pre-processing operation on red, green, blue, and NIR channels separately. They also performed the Gaussian blur operation to remove noise using a [5$\times$5] kernel. To standardise the channels, the values were subtracted by the mean of all channel values and divided by their standard deviation. After that, they normalised and zero-centred the channel values. \textcite{jiang2019deepseedling} applied a Contrast Limited Adaptive Histogram Equalisation algorithm to enhance the image contrast and reduce the image variation due to ambient illumination changes.\par

In the work of \textcite{le2020performances} and  \parencite{bakhshipour2017weed}, all images were segmented  using the Excess Green minus Excess Red Indices (ExG-ExR) method, which effectively removed the background. They also applied opening and closing morphological operations of images and generated contour masks to extract features. On the other hand, \textcite{asad_bais_2019} argued that the Maximum Likelihood Classification technique performed better than thresholding techniques for segmenting the background soil and green plants. According to \textcite{alam2020real}, images captured from the field had many problems (e.g. lack of brightness). It was necessary to apply image pre-processing operations to prepare the data for training. They performed several morphological operations to remove motion blur and light illumination. They also removed the noisy region before applying segmentation operations for separating the background. Threshold-based segmentation techniques had been used to separate the soil and green plants in an image. In the reports of \textcite{espejo2020towards} and \textcite{andrea2017precise}, the RGB channels of the images were normalised  to avoid differences in lighting conditions before removing the background. For vegetation segmentation, Otsu's thresholding was applied, followed by the ExG (Excess Green) vegetation indexing operation. However, \textcite{dyrmann2016plant} used a simple excessive green segmentation technique for removing the background and detecting the green pixels. \textcite{knoll2019real} converted the RGB image to HSV colour space, applied thresholding method and band-pass filtering, and then used binary masking to extract the image's green component.\par

\paragraph{Image Enhancement and Denoising}

\textcite{nkemelu2018deep} investigated the importance of image pre-processing operation by training the CNN model with raw data and processed data. They found that without image pre-processing the model performance decreased. They used Gaussian Blur for smoothing the images and removed the high-frequency content. They then converted the colour of the image to HSV space. Using a morphological erosion with an 11$\times$11 structuring kernel, they subtracted the background soil and produced foreground seedling images. \textcite{lottes2018fully} reported that image pre-processing improved the generalisation capabilities of a classification system. They applied [5$\times$5] Gaussian Kernel to remove noise and to normalise the data. They also zero-centred the pixel values of the image. The study of \textcite{sarvini2019performance} used the Gaussian and median filter to remove Gaussian noise and Salt and Pepper noise respectively. \textcite{tang2017weed} also normalised the data to maintain zero-mean and unit variance. Besides, they applied Principal Component Analysis and Zero-phase Component Analysis data whitening for eliminating the correlation among the data. \par

\textcite{wang2020semantic} evaluated the performance of the DL model based on the input representation of images. They applied many image pre-processing operations, such as histogram equalisation, automatic adjustment of the contrast of images and deep photo enhancement. They also used several vegetation indices including ExG, Excess Red, ExG-ExR, NDVI, Normalised Difference Index, Colour Index of Vegetation, Vegetative Index, and Modified Excess Green Index and Combined Indices. \textcite{liang2019low} split the collected data into blocks which contained multiple plants. The blocks were then divided into sub-images with a single plant in them. After that, the histogram equalisation operation was performed to enhance the contrast of the sub-images.\par

\textcite{rist2019weed} applied orthorectification and radiometric corrections operation to process the satellite image. They then normalised the pixel values of each band. After that, the large satellite image was split into 2138 samples of pixel size 128$\times$128. \par

\subsection{Training Data Generation}

To enlarge the size of the training data, in several related studies data augmentation was applied. It is a very useful technique when the dataset is not large enough \parencite{umamaheswari2020encoder}. If there is a little variation \parencite{sarvini2019performance} or class imbalance \parencite{bah2018deep} among the images of the dataset, the image augmentation techniques are helpful. \textcite{wang2020semantic} applied an augmentation to the dataset to determine the generalisation capability of their proposed approach. Table \ref{tab:data_augmentation} shows different types of data augmentation used in the relevant studies.\par

\FloatBarrier
\begin{table}[bt!]
\centering
\scriptsize
\setstretch{1.0}
\caption{Different types of data augmentation techniques used in the relevant studies}
\label{tab:data_augmentation}
\begin{tabular}{|p{.1\textwidth}|p{.3\textwidth}|p{.5\textwidth}|}
\hline
\multicolumn{1}{|c|}{\textbf{\begin{tabular}[c]{@{}c@{}}Image \\ Augmentation\\ Technique\end{tabular}}} & \multicolumn{1}{c|}{\textbf{Description}} & \multicolumn{1}{c|}{\textbf{Reference}} \\ \hline
Rotation & Rotate the image to the right or   left on an axis between 1$^{\circ}$ and 359$^{\circ}$ \parencite{shorten2019survey} & \parencite{bah2018deep,le2020performances,sarvini2019performance,espejo2020towards,gao2020deep,adhikari2019learning,farooq2018analysis,andrea2017precise,dyrmann2016plant,brilhador2019classification,binguitchacrops,zhang2018broad} \\ \hline

Scaling & Use zooming in/out to resize the image \parencite{kumar_2019}. & \parencite{asad_bais_2019,gao2020deep,adhikari2019learning,brilhador2019classification,binguitchacrops} \\ \hline

Shearing & Shift one part of the image to a direction and the other part to the opposite direction \parencite{shorten2019survey}. & \parencite{asad_bais_2019,le2020performances,gao2020deep,brilhador2019classification, zhang2018broad}  \\ \hline

Flipping & Flip the image horizontally or vertically \parencite{kumar_2019}. & \parencite{asad_bais_2019,abdalla2019fine,sarvini2019performance,gao2020deep,adhikari2019learning,dyrmann2016plant,chechlinski2019system,brilhador2019classification,binguitchacrops, zhang2018broad,petrich2019detection} \\ \hline

Gamma   Correction & Encode and decode the luminance   values of an image \parencite{brasseur}. & \parencite{wang2020semantic} \\ \hline

Colour   Space & Isolating a single colour   channel, increase or decrease the brightness of the image, changing the   intensity values in the histograms \parencite{shorten2019survey}. & \parencite{bah2018deep,wang2020semantic,asad_bais_2019,espejo2020towards,adhikari2019learning,chechlinski2019system,petrich2019detection} \\ \hline

Colour Space Transformations & Increase or decrease the pixel values by a constant value and restricting pixel values to a certain min or max value \parencite{shorten2019survey}. & \parencite{le2020performances,sarvini2019performance,binguitchacrops,petrich2019detection} \\ \hline

Noise Injection & Injecting a matrix of random values to the image matrix. For example: Salt-Pepper noise, Gaussian noise   etc \parencite{shorten2019survey}. & \parencite{sarvini2019performance,espejo2020towards,petrich2019detection} \\ \hline

Kernel filtering  & Sharpening or blurring the image \parencite{shorten2019survey}.  & \parencite{bah2018deep,asad_bais_2019,espejo2020towards,petrich2019detection} \\ \hline

Cropping & Remove a certain portion of an image \parencite{takahashi2018ricap}. Usually this is done at random in case of data augmentation \parencite{shorten2019survey}. & \parencite{asad_bais_2019,adhikari2019learning,farooq2018analysis,petrich2019detection} \\ \hline

Translation & Shift the position of all the image pixels \parencite{huang2018auggan}. & \parencite{asad_bais_2019,abdalla2019fine,brilhador2019classification} \\ \hline

\end{tabular}
\end{table}
\FloatBarrier

As shown in Table \ref{tab:data_augmentation}, it is observed that in most of the studies, different geometric transformation operation were applied to the data. Use of colour augmentation can be helpful to train a model for developing a real-time classification system. This is because the colour of the object varies depending on the lighting condition and motion of the sensors.

Image data that are not collected from the real environments and created artificially or programmatically are known as synthetic data or images \parencite{viraf_2020}. It is not always possible to manage a large amount of labelled data to train a model. In these cases, the use of synthetic data is an excellent alternative to use together with the real data. Several research studies show that artificial data might have a significant change in classifying images \parencite{andreini2020image}. In weed detection using DL approaches, synthetic data generation  technique is not applied very often. \textcite{rasti_ahmad_samiei_belin_rousseau_2019} used synthetically generated images to train the model and achieved a good classification accuracy while testing on a real dataset.\par

On the other hand, \textcite{pearlstein2016convolutional} created complex occlusion of crops and weeds and generated variation in leaf size, colour, and orientation by producing synthetic data. To minimise human effort for annotating data, \textcite{di2017automatic} generated synthetic data to train the model. For that purpose, they used a generic kinematic model of a leaf prototype to generate a single leaf of different plant species and then meshed that leaf to the artificial plant. Finally, they placed the plant in a virtual crop field for collecting the data without any extra effort for annotation. 

\textcite{skovsen2019grassclover} generated a 8000 synthetic dataset for labelling a real dataset. To create artificial data, they cropped out different parts of the plant, randomly selected any background from the real data, applied image processing (e.g. rotation, scaling, etc.), and added an artificial shadow using a Gaussian filter.\par 

\subsection{Data Labelling}

The majority of the reviewed publications used manually annotated data labelled by experts for training the deep learning model in a supervised manner. The researchers applied different annotations, such as bounding boxes annotation, pixel-level annotation, image-level annotation, polygon annotation, and synthetic labelling, based on the research need. Table \ref{table:image_annotation_technique} shows different image annotation approaches used for weed detection. However, \textcite{jiang2020cnn} applied a semi-supervised method to label the images; they used a few labelled images to annotate the unlabelled data. On the other hand, \textcite{dos2019unsupervised} proposed a semi-automatic labelling approach. Unlike semi-supervised data annotation, they did not use any manually labelled data, but applied the clustering method to label the data. First, they divided the data into different clusters according to their features and then labelled the clusters. Similar techniques were used by \textcite{hall_dayoub_perez_mccool_2018}. \textcite{yu2019weed} separated the collected images into two parts; one with positive images that contained weeds, and the other of negative images without weeds. \textcite{lam2020open} proposed an object-based approach to generate labelled data.\par

\FloatBarrier
\begin{table}[bt!]
\centering
\scriptsize
\caption{Different image annotation techniques used for weed detection using deep learning}
\label{table:image_annotation_technique}
\begin{tabular}{|L{0.02\textwidth}|L{0.1\textwidth}|p{0.2\textwidth}|p{0.5\textwidth}|}
\hline
\multicolumn{2}{|c|}{\textbf{\makecell{Type of Image \\ Annotation}}} & \multicolumn{1}{c|}{\textbf{Description}} & \multicolumn{1}{c|}{\textbf{Reference}}                                                   \\ \hline

\multicolumn{2}{|c|}{Pixel Level Annotation}  & Label each pixel whether it belongs to crop or weed in the image. & \textcite{ishak2007weed,bini2020machine, liakos2018machine,huang2020deep, sa2018weedmap, zhang2020weed, kounalakis2019deep, hall_dayoub_perez_mccool_2018, abdalla2019fine, andrea2017precise, fawakherji2019crop,chechlinski2019system,brilhador2019classification,adhikari2019learning,farooq2019multi,bosilj2020transfer,asad_bais_2019,knoll2019real,umamaheswari2020encoder,lottes2018fully,huang2018fully,huang2018accurate,milioto2017real, lottes2020robust, rist2019weed, patidar2020weed, osorio2020deep,ramirez2020deep,lam2020open,skovsen2019grassclover}  \\ \hline

\multirow{2}{*}[-1em]{\begin{sideways}\makecell[c]{Region Level Annotation}\end{sideways}}& Bounding Boxes Annotation & There may be a mixture of weeds and crops in a single image. Using a bounding box the crops and weeds are labelled in the image. & \textcite{huang2018semantic, sa2017weednet, bah2018deep, kounalakis2018robotic, ma2019fully, nkemelu2018deep, farooq2018analysis, sharpe2019detection, jiang2019deepseedling, binguitchacrops,gao2020deep,veeranampalayam2020comparison,partel2019development,rasti_ahmad_samiei_belin_rousseau_2019,valente2019detecting,dyrmann2017roboweedsupport, petrich2019detection, sharpe2020goosegrass}  \\ \cline{2-4}
& Polygon Annotation & This is used for semantic segmentation to detect irregular shaped object. It outlines the region of interest with arbitrary number of sides. & \textcite{patidar2020weed} 
\tabularnewline \hline

\multicolumn{2}{|c|}{Image Level Annotation} & Uses separate image for weeds and crops to train the model.  & \textcite{yu2019deep, czymmek2019vision, wang2020semantic,zhang2018broad,partel2019smart, le2020performances, suh2018transfer, alam2020real,sarvini2019performance,yu2019weed,teimouri2018weed,tang2017weed,espejo2020towards, farooq2018weed,yan2020classification, liang2019low,jiang2020cnn,olsen2019deepweeds, dos2017weed, trong2020late}\\ \hline
\multicolumn{2}{|c|}{Synthetic Labelling}  & For training the model use generated and labelled data. & \parencite{pearlstein2016convolutional, di2017automatic}                                                  \\ \hline
\end{tabular}
\end{table}
\FloatBarrier

As summarised in Table \ref{table:image_annotation_technique}, commonly used annotation techniques are bounding boxes, pixel-wise labelling and image level annotation. However, plants are irregular in shape: by using polygon annotation, the images of crops and weeds can be separated accurately. Synthetic labelling approaches can minimise labelling costs and help to generate large annotated datasets. 

\section{Detection Approaches} \label{Detection_Approach}

Studies in this area apply two broad approaches for detecting, localising, and classifying weeds in crops: i) localise every plant in an image and classify that image either as a crop or as a weed; ii) map the density of weeds in the field. To detect weeds in crops, the concept of \enquote{row planting} has been used. In some of these studies, there are  further classification steps of the weed species.\par

\subsection{Plant-based Classification}

To develop a weed management system, a major step is to classify every plant as weed or crop plant \parencite{lottes2018joint}. The first problem is to detect weeds, followed by localisation and finally, classification. This approach is useful for real-time weed management techniques. For instance, \textcite{raja2020real} developed a real-time weeding system where a robotic machine detected the weeds and used a knife to remove them. In this case, it was necessary to label individual plants, whether as a weed or crop plant. In traditional farming approaches, farmers usually apply a uniform amount of herbicide over the whole crop in a field. A machine needs to identify individual crop plants and weeds to apply automatic selective spraying techniques. Besides, identifying the weed species is also important to apply specific treatments \parencite{lottes2020robust}. We have found that this approach has been used in most of the studies reported.\par

\subsection{Weed Mapping}

Mapping weed density can also be helpful for site-specific weed management and can lead to a reduction in the use of herbicides. \textcite{huang2020deep} used the DL technique to map the density of weeds in a rice field. An appropriate amount of herbicides can be applied to a specific site based on the density map. The work in \textcite{abdalla2019fine} segmented the images and detected the weed presence in the region of that image. Using a deep learning approach, \textcite{huang2018semantic} generated a weed distribution map of the field. In addition, some researchers argued that weed mapping helps to monitor the conditions of the field automatically \parencite{sa2017weednet,sa2018weedmap}. Farmers can monitor the distribution and spread of weeds, and can take action accordingly.\par

\section{Learning Methods} \label{learning_mehod}

\subsection{Supervised Learning}

Supervised learning occurs when the datasets for training and validation are labelled. The dataset passed in the DL model as input contains the image along with the corresponding labels. That means, in supervised training, the model learns how to create a map from a given input to a particular output based on the labelled dataset. Supervised learning is popular to solve classification and regression problems \parencite{caruana2006empirical}. In most of the related research the supervised learning approach was used to train the DL models. Section \ref{Deep_Learning_Architecture} presents a detail description of those DL architectures. \par

\subsection{Unsupervised Learning}

Unsupervised learning occurs when the training set is not labelled. The dataset passed as input in the unsupervised model has no corresponding annotation. The models attempt to learn the structure of the data and extract distinguishable information or features from data. Using this process, the model becomes able to map the input to the particular output. From this, the objects in the whole dataset will be divided into separate groups or clusters. The features of the objects in a cluster are similar and differ from other clusters. This is how unsupervised learning can classify objects of a dataset into separate categories. Clustering is one of the applications of unsupervised learning \parencite{barlow1989unsupervised}.\par

Most of the relevant studies used a supervised learning approach to detect and classify weeds in crops automatically. However, \textcite{dos2019unsupervised} proposed unsupervised clustering algorithms with a semi-automatic data labelling approach in their research. They applied two clustering methods- Joint Unsupervised Learning (JULE) and Deep Clustering for Unsupervised Learning of Visual Features algorithms (DeepCluster). They developed the models using AlexNet \parencite{krizhevsky2012imagenet} and VGG-16 \parencite{simonyan2014very} architecture and initialised with pre-trained weights. They achieved a promising result (accuracy 97\%) in classifying weeds in crops and reduce the cost of manual data labelling.\par

\textcite{tang2017weed} applied an unsupervised K-means clustering algorithm as a pre-training process and generate a feature dictionary. They then used those features to initialise the weights of the CNN model. They claimed that it can improve generalisation ability in feature extraction and resolve the unstable identification problem. The proposed approach shows better accuracy than SVM, Back Propagation neural network, and even CNN with randomly initialised weights.\par

\subsection{Semi-supervised Learning}

Semi-supervised learning takes the middle ground between supervised and unsupervised learning \parencite{lee2013pseudo}. A few researchers used  Graph Convolutional Network (GCN) \parencite{kipf2016semi} in their research, which is a semi-supervised model. The major difference between CNN and GCN is the structure of input data. CNN is for regular structured data, whereas GCN uses graph data structure \parencite{mayachita_2020}. We discuss the use of GCN in the related work in Section \ref{GCN_Network}.  \par

\section{Deep Learning Architecture} \label{Deep_Learning_Architecture}

Our analysis shows that the related studies apply different DL architectures to classify the weeds in crop plants based on the dataset and research goal. Most researchers compared their proposed models either with other DL architecture or with traditional machine learning approaches. Table \ref{tab:different_DL_approach} shows an overview of different DL approach used in weed detection. A CNN model generally consists of two basic parts- feature extraction and classification \parencite{khoshdeli2017detection}. In related research, some researchers applied CNN models using various permutation of feature extraction and classification layers. However, in most cases, they preferred to use state-of-art CNN models like VGGNet \parencite{simonyan2014very}, ResNet (deep Residual Network) \parencite{he2016deep}, AlexNet \parencite{krizhevsky2012imagenet}, InceptionNet \parencite{szegedy2015going}, and many more. Fully Convolutional Networks (FCNs) like SegNet \parencite{badrinarayanan2017segnet} and U-Net \parencite{ronneberger2015u} were also used in several studies.\par

\subsection{Convolutional Neural Network (CNN)}


\subsubsection{Pre-trained Network}

\textcite{suh2018transfer} applied six well known CNN models namely AlexNet, VGG-19, GoogLeNet, ResNet-50, ResNet-101 and Inception-v3. They evaluated the network performance based on the transfer learning approach and found that pre-trained weights had a significant influence on training the model. They obtained the highest classification accuracy (98.7\%) using the VGG-19 model, but it took the longest classification time. Considering that, the AlexNet model worked best for detecting volunteer potato plants in sugar beet according to their experimental setup. Even under varying light conditions, the model could classify plants with an accuracy of about 97\%. The study of \textcite{dos2017weed} also supported that. They compared the classification accuracy of AlexNet with SVM, Adaboost – C4.5, and the Random Forest model. The AlexNet architecture performed better than other models in discriminating soybean crop, soil, grass, and broadleaf weeds. Similarly, \textcite{valente2019detecting} reported that the AlexNet model with pre-trained weights showed excellent performance for detecting Rumex in grasslands. They also showed that by increasing heterogeneous characteristics of the input image might improve the model accuracy (90\%). However, \textcite{lam2020open} argued that to detect Rumex in grassland the VGG-16 model performs well with an accuracy of 92.1\%.\par

\textcite{teimouri2018weed} demonstrated that, although ImageNet dataset does not contain the images of different plant species, the pre-trained weights of the dataset could still help to reduce the number of training iterations. They fine-tuned Inception-v3 architecture for classifying eighteen weed species and determining growth stages based on the number of leaves. The model achieved the classification accuracy of 46\% to 78\% and showed an average accuracy of 70\% while counting the leaves. However, \textcite{olsen2019deepweeds} differed from them. They developed a multi-class weed image dataset consisting of eight nationally significant weed species. The dataset contains 17,509 annotated images collected from different locations of northern Australia. They also applied the pre-trained Inception-v3 model along with ResNet-50 to classify the weed species (source code is available here: \url{https://github.com/AlexOlsen/DeepWeeds}). The average classification accuracy of ResNet-50 (95.7\%) was a little higher than Inception-v3 (95.1\%). \textcite{bah2018deep} also used ResNet with pre-trained weights as they found it more useful to detect weeds.\par

According to \textcite{yu2019deep}, Deep Convolutional Neural Network (DCNN) can perform well in detecting different species of weeds in bermudagrass. They used three pre-trained (from ImageNet dataset and KITTI dataset \parencite{geiger2013vision}) models including VGGNet, GoogLeNet and DetectNet. In another study, they added AlexNet architecture with the previous models for detecting weeds in perennial ryegrass  \parencite{yu2019weed}. Though all the models performed well, DetectNet exhibited a bit higher F1 score of $\geq$0.99. On the other hand, \textcite{sharpe2019detection} evaluated the performance of VGGNet, GoogLeNet, and DetectNet architecture using two variations of images (i.e., whole and cropped images). They also agreed that the DetectNet model could detect and classify weed in strawberry plants more accurately using cropped sub-images. They suggested that the most visible and prevalent part of the plant should be annotated rather than labelling the whole plant in the image.\par

\textcite{le2020performances} proposed a model namely Filtered LBP (Local Binary Patterns) with Contour Mask and coefficient k (k-FLBPCM). They compared the model with VGG-16, VGG-19, ResNet-50, and Inception-v3 architecture. The k-FLBPCM method effectively classified barley, canola and wild radish with an accuracy of approximately 99\%, which was better than other CNN models (source code is available here: \url{https://github.com/vinguyenle/k-FLBPCM-method}). The network was trained using pre-trained weights from ImageNet dataset. \par

\textcite{andrea2017precise} compared the performance of LeNET \parencite{lecun1989backpropagation}, AlexNet, cNET \parencite{gabor1996automated}, and sNET \parencite{qin2019thundernet} in their research. They found that cNET was better in classifying maize crop plants and their weeds. They further compared the performance of the original cNET architecture with the reduced number of filter layers (16 filter layers). The result reported that with pre-processed images, 16 filter layers were adequate to classify the crops and weeds. Besides, it made the model 2.5 times faster than its typical architecture and helped to detect weeds in real-time.\par

\textcite{partel2019smart} analysed the performance of Faster R-CNN \parencite{ren2015faster}, YOLO-v3 \parencite{redmon2018yolov3}, ResNet-50, ResNet-101, and Darknet-53 \parencite{redmon} models to develop a smart sprayer for controlling weed in real-time. Based on precision and recall value, ResNet-50 model performed better than others. In contrast, \textcite{binguitchacrops} applied the ResNet-101 model. They demonstrated that the size of the input image could affect the performance of ResNet-101 architecture. They used three different pixel sizes (i.e., 128px, 224px, and 360px) for their experiment and reported that model accuracy gets better by increasing the pixel size of the input image.\par

\textcite{trong2020late} proposed a multi-modal DL approach for classifying species of weeds. In this approach, they trained five pre-trained DL models, including NASNet, ResNet, Inception–ResNet, MobileNet, and VGGNet independently. The Bayesian conditional probability-based technique and priority weight scoring method were used to calculate the score vector of models. The model with better scores has a higher priority on determining the classes of species. To classify weed species, they summed up the probability vectors generated by the softmax layer of each model and the species with the highest probability value was determined. According to the experimental results, they argued that the performance of this approach was better than a single DL model.\par       

\subsubsection{Training from Scratch}

\textcite{dyrmann2016plant} argued that, a CNN model initialised with pre-trained weights which was not trained on any plant images would not work well. They therefore built a new architecture using a combination of convolutional layers, batch normalisation, activation functions, max-pooling layers, fully connected layers, and residual layers according to their need. The model was used to classify twenty-two plant species, and they achieved a classification accuracy ranging from 33\% to 98\%.\par

\textcite{milioto2017real} built a CNN model for blob wise discrimination of crops and weeds. They used multi-spectral images to train the model. They investigated different combinations of convolutional layers and fully connected layers to explore an optimised, light-weight and over-fitting problem-free model. Finally, using three convolutional layers and two fully connected layers, they obtained a better result. They stated that this approach did not have any geometric priors like planting the crops in rows. \textcite{farooq2018analysis} claimed in their research that the classification accuracy of the CNN model depended on the number of the hyperspectral band and the resolution of the image patch. They also built a CNN model using a combination of convolutional, nonlinear transformation, pooling and dropout layers. In further research, they proved that a CNN model trained with a higher number of bands could classify images more accurately than HoG (Histogram of oriented Gradients) based method \parencite{farooq2018weed}.\par 

\textcite{nkemelu2018deep} compared CNN's performance with SVM (61.47\%) and K-Nearest Neighbour (KNN) algorithm (56.84\%) and found that CNN could distinguish crop plants from weeds better. They used six convolutional layers and three fully connected layers in the CNN architecture to achieve the accuracy of 92.6\%. They also evaluated the accuracy of CNN using the original images and the pre-processed images. The experimental results suggested that classification accuracy improved by using pre-processed images. \textcite{sarvini2019performance} agreed that CNN offers better accuracy than SVM and ANN in detecting weeds in crop plants because of its deep learning ability. \textcite{liang2019low} employed a CNN architecture that consists of three convolutional, three pooling, four Dropout layers, and a fully connected layer for developing a low-cost weed recognition system. Their experiment also proved that the performance of the CNN model in classification was better than the HoG and LBP methods. \textcite{zhang2018broad} also demonstrated that the CNN model was better than SVM for detecting broad-leaf weeds in pastures. They used a CNN model with six convolutional layers and three fully connected classification layers. The model could recognise weeds with an accuracy of 96.88\%, where SVM achieved maximum accuracy of 89.4\%.\par

\textcite{pearlstein2016convolutional} used synthetic data to train their CNN model and evaluated it on real data. They built a CNN model with five convolutional layers and two fully connected layers. The results showed that CNN could classify crop plants and weeds very well from natural images and with multiple occlusion. Although \textcite{rasti_ahmad_samiei_belin_rousseau_2019} applied the same architecture in their research, they argued that the Scatter Transform method achieved better accuracy with a small dataset than the CNN architecture. They compared several machine learning approaches like Scatter Transform, LBP, GLCM, Gabor filter with the CNN model. They also used synthetic data for training and evaluated the models' performance on real field images.\par

\subsection{Region Proposal Networks (RPN)}

Based on the tiny YOLO-v3 \parencite{yi2019improved} framework, \textcite{gao2020deep} proposed a DL model which speeds up the inference time of classification (source code is available here: \url{https://drive.google.com/file/d/1-E_b_5oqQgAK2IkzpTf6E3X1OPm0pjqy/view?usp=sharing}). They added two extra convolutional layers to the original model for better feature fusion and also reduced the number of detection scales to two. They trained the model with both synthetic data and real data. Although YOLO-v3 archived better classification accuracy in the experiments, they recommended the tiny YOLO-v3 model for real-time application. \textcite{sharpe2020goosegrass} also used tiny YOLO-v3 model to detect goosegrass in strawberry and tomato plants.\par

YOLO-v3 and tiny YOLO-v3 models were also employed in a research by \textcite{partel2019development}. The aim was to find a low-cost, smart weed management system. They applied the models on two machines with different hardware configurations. Their paper reported that YOLO-v3 showed good performance when tested on powerful and expensive computers, but the processing speed decreased if executed on a lower power computer. From their experiments, they came to the conclusion that to save the hardware cost, the tiny YOLO-v3 model was better. \textcite{zhang2018broad} also preferred to use tiny YOLO-v3 instead of YOLO-v3, because it was a lightweight method and took less time and resources to classify objects. In contrast, \textcite{czymmek2019vision} proposed to use YOLO-v3 with a relatively larger input image size (832 $\times$ 832 pixels). They argued that the model performed better in their research with a small dataset. They agreed that tiny YOLO-v3 or Fast YOLO-v3 could improve the detection speed, but there was a need to compromise with the model accuracy.\par

\textcite{veeranampalayam2020comparison} trained and evaluated the performance of a pre-trained Faster R-CNN and SSD (Single Shot Detector) \parencite{liu2016ssd} object detection models to detect late-season weed in soybean fields. Moreover, they compared these object detection models with patch-based CNN model. The result showed that Faster R-CNN performed better in terms of weed detection accuracy and inference speed. \textcite{jiang2019deepseedling} proposed the Faster R-CNN model to detect the weeds and crop plants and to count the number of seedlings from the video frames. They used Inception-ResNet-v2 architecture as the feature extractor. On the other hand, by applying the Mask R-CNN model on \enquote{Plant Seedlings Dataset} \textcite{patidar2020weed} achieved more than 98\% classification accuracy. They argued that Mask R-CNN detected plant species more accurately with less training time than FCN. \par

\textcite{osorio2020deep} compared two RPN models, namely YOLO-v3 and Mask R-CNN with SVM. The classification accuracy of RPN architectures was 94\%, whereas SVM achieved 88\%. However, they reported that as SVM required less processing capacity, it could be used for IoT based solution.\par

\subsection{Fully Convolutional Networks (FCN)}

Unlike CNN, FCN replaces all the fully connected layers with convolutional layers and uses a transposed convolution layer to reconstruct the image with the same size as the input. It helps to predict the output by making a one-to-one correspondence with the input image in the spatial dimension \parencite{shelhamer2017fully, huang2020deep}.

\textcite{huang2018fully} compared the performance of AlexNet, VGGNet, and GoogLeNet as the base model for FCN architecture. VGGNet achieved the best accuracy among those. They further compared the model with patch-based CNN and pixel-based CNN architectures. The result showed that the VGG-16  based FCN model achieved the highest classification accuracy. On the other hand, \textcite{huang2018semantic} applied ResNet-101 and VGG-16 as a baseline model of FCN for segmentation. They also compared the performance of the FCN models with a pixel-based SVM model. In their case, ResNet-101 based FCN architecture performed better. \textcite{asad_bais_2019} compared two FCN architecture for detecting weeds in canola fields, i.e., SegNet and U-Net. They used VGG-16 and ResNet-50 as the encoder block in both the models. The SegNet with ResNet-50 as the base model achieved the highest accuracy.\par  

According to \textcite{ma2019fully}, SegNet (accuracy 92.7\%) architecture was better than traditional FCN (accuracy 89.5\%) and U-Net (accuracy 70.8\%) for weed image segmentation when classifying rice plants and weeds in the paddy field. The study of \textcite{abdalla2019fine} reported that the accuracy of image segmentation depended on the size of the dataset. That is why it is difficult to train a model from scratch. To address this problem, they applied transfer learning and real-time data augmentation to train the model. In their experiment, they used VGG-16 based SegNet architecture. They applied three different transfer learning approaches for VGG-16. Moreover, the performance of the model was compared with the VGG-19 based architecture. The VGG-16 based SegNet achieved the highest accuracy of 96\% when they used pre-trained weights only for feature extraction and the shallow machine learning classifier (i.e., SVM) for segmentation. \textcite{sa2017weednet} also applied SegNet with the pre-trained VGG-16 as the base model (source code is available here: \url{https://github.com/inkyusa/weedNet}). They trained the model by varying the number of channels in the input images. They then compared the inference speed and accuracy of different arrangements by deploying the model on an embedded GPU system, which was carried out by a small micro aerial vehicle (MAV). \textcite{umamaheswari2020encoder} compared the performance of SegNet-512 and SegNet-256 encoder-decoder architectures for semantic segmentation of weeds in crop plants. The experiment proved that SegNet-512 was better for classification. In the study of \textcite{di2017automatic}, the SegNet model was trained using synthetic data, and the performance was evaluated on a real crop and weed dataset.\par

\textcite{fawakherji2019crop} proposed U-Net architecture using VGG-16 as an encoder for semantic segmentation. They also applied a VGG-16 model for classifying the crop plants and weeds. They also trained the model with one dataset containing sunflower crop and evaluated it with two different datasets with carrots and sugar beets crops. In the work of \textcite{rist2019weed}, a ResNet based U-Net model was employed to map the presence of gamba grass in the satellite image. However, \textcite{ramirez2020deep} compared the performance of DeepLab-v3 \parencite{chen2017rethinking} with  SegNet and U-Net model in their research. The results demonstrated that DeepLab-v3 architecture achieved better classification accuracy using class balanced data that has greater spatial context. \par

\textcite{lottes2018fully} also proposed FCN architecture using DenseNet as a baseline model. Their novel approach provided a pixel-wise semantic segmentation of crop plants and weeds. The work of \textcite{lottes2020robust} proposed a task-specific decoder network. As the plants were sown at a regular distance, they trained the model in a way so that the model could learn the spatial plant arrangement from the image sequence. They then fused this sequential feature with the visual features to localise and classify weeds in crop plants. \textcite{dyrmann2017roboweedsupport} used FCN architecture not only for segmentation but also for generating bounding boxes around the plants. They applied pre-trained GoogLeNet architecture as the base model.\par

According to \textcite{wang2020semantic}, changes in the input representation could make a difference in classification performance. They employed the encoder-decoder deep learning network for semantic segmentation of crop and weed plants by initialising the input layers with pre-trained weights. They evaluated the model with different input representation by including NIR information with colour space transformation on the input, which improved crop-weed segmentation and classification accuracy (96\%). \textcite{sa2018weedmap} also evaluated different input representation to train the network. They applied VGG-16 based SegNet architecture for detecting background, crop plants and weeds. The model was evaluated by varying the number of spectral bands and changing the hyper-parameters. The experimental results showed that the model achieved far better accuracy by using nine spectral channels of an image rather than the RGB image.\par

\textcite{huang2018accurate} stated that the original FCN-4s architecture was designed for PASCAL VOC 2011 dataset, which had 1000 classes of objects. However, their dataset had only three categories (i.e., rice, weeds, and others). As a result they reduced the feature maps of the intermediate layers to 2048. They then compared the accuracy and efficiency of the model with original FCN-8s and DeepLab architecture and proved that the modified FCN-4s model performed better. For the same reason, \textcite{bosilj2020transfer} simplified the original architecture of SegNet and named it as SegNet‐Basic. They decreased the number of convolutional layers from 13 to 4.\par

One of the problems with the basic architecture of FCN is that the spatial features can not be recovered properly. The prediction accuracy can be decreased due to this issue. To address this problem, \textcite{huang2020deep} improved the model by adding skip architecture (SA), fully connected conditional random fields and partially connected conditional random fields. They fine-tuned AlexNet, VGGNet, GoogLeNet, and ResNet based FCN. They then compared the performance of different FCNs and Object-based image analysis (OBIA) method. Experimental results reported that the VGGNet-based FCN with proposed improvements achieved the highest accuracy.\par

\textcite{brilhador2019classification} modified the original U-Net architecture for pixel-level classification of crop plants and weeds. They added a convolutional layer with a kernel size of 1$\times$1. For that change, they adjusted the input size of the network. Besides, replacing the ReLU activation functions with the Exponential Linear Unit (ELU), they used adadelta optimiser algorithm instead of the stochastic gradient descent and included dropout layers in between convolutional layers. \textcite{petrich2019detection} also modified the U-Net model to detect one species of weed in grasslands.\par

\subsection{Graph Convolutional Network (GCN)} \label{GCN_Network}

\textcite{hu2020graph} proposed Graph Weeds Net (GWN). GWN is a graph-based deep learning architecture to classify weed species. \textcite{hu2020graph} used ResNet-50 and DenseNet-202 model to learn vertex features with graph convolution layers, vertex-wise dense layers, and the multi-level graph pooling mechanisms included in GWN architecture. Here, an RGB image was represented as a multi-scale graph. The graph-based model with DenseNet-202 architecture achieved the classification accuracy of 98.1\%.\par

\textcite{jiang2020cnn} proposed ResNet-101 based graph convolutional network in their research. They chose GCN, because it was a semi-supervised learning approach. Moreover, the feature relationships were captured using a graph structure. In this model, the label information was shared by neighbouring vertices of the graph, which make the learning more accurate with limited annotated data. They compared the proposed model with AlexNet, VGG-16, and ResNet-101 architecture on four different datasets. The GCN approach achieved 97.80\%, 99.37\%, 98.93\% and 96.51\% classification accuracy for each dataset.\par

\subsection{Hybrid Networks (HN)}

Hybrid architectures are those where the researchers combine the characteristics of two or more DL models. For instance, \textcite{chavan2018agroavnet} proposed the AgroAVNET model, which was a hybrid of AlexNet and VGGNet architecture. They chose VGGNet for setting the depth of filters and used the normalisation concept of AlexNet. They then compared the performance of the AgroAVNET network with the original AlexNet and VGGNet and their different variants. All the parameters were initialised using pre-trained weights except for the third layer of the fully connected layers. They initialised that randomly. The AgroAVNET model outperformed others with a classification accuracy of 98.21\%. However, \textcite{farooq2019multi} adopted the feature concatenation approach in their research. They combined a super pixel-based LBP (SPLBP) method to extract local texture features, CNN for learning the spatial features and SVM for classification. They compared their proposed FCN-SPLBP model with CNN, LBP, FCN, and SPLBP architectures.\par

\textcite{stewart2016end} proposed OverFeat-GoogLeNet architecture by combining the features from LSTM and GoogLeNet model. The model was used to develop a \enquote{Parallelised Weed Detection System} by \textcite{umamaheswari2018weed}. They claimed that this system was robust, scalable and could be applied for real-time weed detection. The classification accuracy of the system was 91.1\%.\par

\textcite{kounalakis2019deep} fine-tuned AlexNet, VGG-F, VGG-16, Inception-v1, ResNet-50, and ResNet-101 model to extract features from the images. They replaced the CNNs' default classifiers with linear classifiers, i.e., SVM and logistic regression. They compared the performance of various SVM and logistic regression classifiers by combining them with CNN models for detecting weeds. They achieved the most balanced result in terms of accuracy and false positive rate by using \enquote{L2-regularised with L2-loss logistic regression model using primal computation} classifier. This classifier performed better while being used with GoogLeNet architecture for detecting weeds in grasslands \parencite{kounalakis2018robotic}. \par

\textcite{espejo2020towards} also replaced CNN's default classifier with traditional ML classifiers including SVM, XGBoost, and Logistic Regression. They initialised Xception, Inception-ResNet, VGGNets, MobileNet, and DenseNet model with pre-trained weights. The experimental result showed that the best performing network was DenseNet model with the SVM classifier. The micro F1 score for the architecture was 99.29\%. This research also reported that with a small dataset, network performance could be enhanced using this approach.\par

\textcite{adhikari2019learning} proposed a fully convolutional encoder-decoder network named as Enhanced Skip Network. The model had multiple VGGNet-like blocks in the encoder and decoder. However, the decoder part had fewer future maps to reduce the computational complexity and memory requirement. Besides, the skip layers, larger convolutional kernels and a multi-scale filter bank were incorporated in the proposed model. The weights were initialised using the transfer learning method. The model performed better than U-Net, FCN8, and DeepLab-v3, Faster R-CNN, and EDNet in identifying weeds in the paddy field. \textcite{chechlinski2019system} combined U-Net architecture, MobileNet-v2 and DenseNet architectures and replaced transposed convolution layers with activation map scaling.\par

\section{Performance Evaluation Metrics} \label{Evaluation_Matrics}

In general, evaluation metrics is the measurement tool to quantify the performance of a classifier. Different metrics are used to evaluate various characteristics of a classifier \parencite{hand2009measuring}. The evaluation metrics can be used either to measure the quality of a classification model \parencite{hand2009measuring} or to compare the performance of the different trained models for selecting the best one \parencite{ozcift2011classifier}. Various metrics were used in related studies based on the research need. The most commonly used metric is classification accuracy (CA) to evaluate the DL model. Many of the authors used multiple metrics to assess the model before drawing any conclusion. Table \ref{tab:evaluation_metrics} lists the evaluation metrics applied in the relevant studies.\par

As Table \ref{tab:evaluation_metrics} shows, it is not easy to compare the related works as different types of evaluation metrics are employed depending on the DL model, the goal of classification, dataset and detection approach. However, the most frequently used evaluation metrics are CA, F1 score and mIoU. In the case of classifying plant species, researchers prefer to use confusion metrics to evaluate the model.\par 

\FloatBarrier
\begin{table}[bt!]
    \scriptsize
    \centering
        \caption{The evaluation metrics applied by different researchers of the related works}
        \label{tab:evaluation_metrics}
    \begin{tabular}{|p{0.03\textwidth}|L{0.25\textwidth}|p{0.6\textwidth}|}

        \hline
        \multicolumn{1}{|c|}{\textbf{No.}} & \multicolumn{1}{c|}{\textbf{Performance Metric}} & \multicolumn{1}{c|}{\textbf{Meaning}} \\    
        \hline
        1.   & Classification Accuracy (CA) & The percentage of correct   prediction among the input. A model is judged based on how high the value is \\ \hline
        
        2.   & True Positive (TP) & How many times the model correctly   predict the actual classes of the object. \\ \hline
        
        3.  & False Positive Rate (FPR) & It is the proportion of negative   cases incorrectly identified as positive cases in the data. \\ \hline
        
        4.  & False Negative Rate (FNR) & The ratio of positive samples that   were incorrectly classified. \\ \hline
        
        5.  & Specificity (S) & The fraction of True Negative from   the sum of False Positive and True Negative. \\ \hline
        
        6.  & Mean Pixel Accuracy (MPA) & It is the average of ration of the   correctly classified pixels among all pixels of the images in the dataset. It   is used to evaluate the model for semantic segmentation. \\ \hline
        
        7. & Precision (P)    & The fraction of correct prediction   (True Positive) from the total number of relevant result (Sum of True   Positive and False Positive).  It helps   when the value of False Positives are high. \\ \hline
        
        8.    & Mean Average Precision (mAP) & It is the mean of average   precision over all the classes of an object in the data. \\ \hline
        
        9.   & Recall & The fraction of True Positive from   the sum of True Positive and False Negative. It helps when the value of False Negatives are high. \\ \hline

        10.   & F1 Score (F1) & The harmonic mean of precision and   recall. \\ \hline
        
        11.   & Confusion Matrix (CM) & It is the summary of the number of   correct and incorrect prediction made by a model. It helps to visualise not   only the errors made by the model but also the types of error in predicting   the class of object. \\ \hline
        
        12.   & Intersection over Union (IoU) & It is the ratio of the overlapping   area of ground truth (the hand labelled bounding boxes from the testing   dataset) and predicted area (predicted bounding boxes from the model) to the   total area. \\ \hline
        
        13.   & Mean Intersection over Union   (mIoU) & It is average IoU over all the   classes of an object in the dataset. \\ \hline
        
        14.   & Frequency Weighted Intersection   over Union (FWIoU) & It is the weighted average of IoUs   based on pixel classes. \\ \hline
        
        15.  & Mean Square Error (MSE) & It is  the mean of all the squared errors between   the predicted and actual target class \\ \hline
        
        16.  & Root Mean Square Error (RMSE) & It is the standard deviation of   the difference between the predicted value and observed values. \\ \hline
        
        17.  & Mean Absolute Error (MAE) & It is the mean of the absolute   values of each prediction error on all instances of the test dataset. \\ \hline
        
        18.  & R2 & It is the squared correlation   between the observed and the predicted outcome by the model. \\ \hline
        
        19.  & K-fold Cross Validation & The dataset is divided into K   number of parts and each of the parts is used as testing dataset. \\ \hline
        
        20.  & Receiver Operating Characteristic   (ROC) curve      & The true positive rate is plotted   in function of the false positive rate for different cut-off points of a   parameter. \\ \hline
        
        21.  & Kappa Coefficient & Measures the degree of agreement between the true values and the   predicted values \\ \hline
        
        22.  & Matthews correlation coefficient   (MCC) & A correlation coefficient between   the observed and predicted binary classifications \\ \hline
        
        23.  & Dice Similarity Coefficient (DSC) & It is a measure of spatial overlap   between two sets of pixels. \\ \hline
        
    \end{tabular}
\end{table}
\FloatBarrier

In addition to the evaluation metrics provided in Table \ref{tab:evaluation_metrics}, \textcite{milioto2017real} justified their model based on run-time. This was because, to develop a real-time weeds and crop plants classifier, it is important to identify the class of a plant as quickly as possible. They showed how quickly their model could detect a plant in an image. Similarly, \textcite{suh2018transfer} calculated the classification accuracy of their model along with the time required to train and identify classes of plants, as they intended to develop a real-time classifier.  \textcite{ma2019fully} also used run-time for justifying the model performance. They found that, by increasing the patch size of the input images, it was possible to reduce the time required to train the model. Another research method used inference time to compare different DL architecture \parencite{huang2018fully}. \textcite{dos2017weed} evaluated the CNN model not only based on time but also in terms of the memory consumed by the model during training. They argued that though the CNN architecture achieved higher accuracy than other machine learning model, it required more time and memory to train the model. \textcite{andrea2017precise} showed that reducing the number of layers of the DL model could make it faster in detecting and identifying the crop and weed plants. They also used processing time as an evaluation criterion while choosing the CNN architecture.\par

\section{Discussion} \label{discussion}

It is evident that the DL model offers high performance in the area of weed detection and classification in crops. In this paper, we have provided an overview of the current status of the area of the automatic weed detection technique. In most relevant studies, the preferred method to acquire data was using a digital camera mounted on a ground vehicle to collect RGB images. A few research studies collected multi-spectral or hyper-spectral data. To prepare the dataset for training, different image processing techniques were used to resize the images, background and noise removing and image enhancement. The datasets were generally annotated using bounding boxes, pixel-wise and image level annotation approaches. For training the model, supervised learning approaches are applied by the researchers. They employ different DL techniques to find a better weed detection model. Detection accuracy is given as the most important parameter to evaluate the performance of the model.\par

Nevertheless, there is still room for improvements in this area. Use of emerging technologies can help to improve the accuracy and speed of automatic weed detection systems. As crop and weed plants have many similarities, the use of other spectral indices can improve the performance. \par

However, there is a lack of large  datasets for crops and weeds. It is necessary to construct a large benchmark dataset by capturing a variety of crops/weeds from different geographical locations, weather conditions and at various growth stages of crops and weeds. At the same time, it will be expensive to annotate these large datasets. Semi-supervised \parencite{chapelle2009semi, zhang2019active} or weakly supervised \parencite{zhou2018brief, durand2017wildcat} approaches could be employed to address this problem.\par

Moreover, Generative Adversarial Network (GAN) \parencite{ledig2017photo} or other synthetic data generation techniques can contribute to creating a large dataset. Random point generation and polygon labelling can further improve the precision of automatic weed detection systems. DL is evolving very fast, and new state-of-art techniques are being proposed. In addition to developing new solutions, researchers can enhance and  apply those methods in the area of weed detection. They can also consider using weakly supervised, self-supervised or unsupervised approaches like multiple instance learning, few-shot or zero-shot learning as a means for synthetic data generation.\par

Furthermore, most datasets mentioned in this paper exhibit class imbalance, which may create biases and lead to over-fitting of the model. Future research needs to address the problem. This can be achieved via the use of appropriate data redistribution approaches, cost-sensitive learning approaches \parencite{khan2017cost}, or class balancing classifiers \parencite{taherkhani2020adaboost,bi2018empirical}. \par  

To summarise, the primary objective of developing automatic weed detection system is to provide a weed management technique that will minimise cost and maximise crop yields. To do so, researchers need to come up with a system that can be deployed on devices with a lower computational requirement and can detect weeds accurately in real-time.\par   

\section{Conclusion} \label{conclusion}

This study provides a comprehensive survey of the deep learning-based research in detecting and classifying weed species in value crops. A total of 70 relevant papers have been examined based on data acquisition, dataset preparation, detection and classification methods and model evaluation process. Publicly available datasets in the related field are also highlighted for prospective researchers. In this article, we provide a taxonomy of the research studies in this area and summarise the approaches of detecting weeds (Table \ref{tab:different_DL_approach}). It was found that most of the studies applied supervised learning techniques using state-of-art deep learning models and they can achieve better performance and classification accuracy by fine-tuning pre-trained models on any plant dataset. The results also show that the experiments already have achieved very high accuracy when a sufficient amount of labelled data of each class is available for training the models. However, the existing research only achieved high accuracy in a limited experiment setup, e.g., on small datasets of a select number of crops and weeds species. Computational speed in the recognition process is another limiting factor for deployment on real-time fast-moving herbicide spraying vehicles. An important future direction would be to investigate highly efficient detection techniques using very large datasets with a variety of crop and weed species so that one single model can be used across any weed-crop setting as needed. Other potential future research directions include the need for large generalised datasets, tailored machine learning models in weed-crop settings, addressing the class imbalance problems, identifying the growth stage of the weeds, as well as thorough field trials for commercial deployments. \par

\cleardoublepage

\cleardoublepage
\printbibliography

\end{document}